\documentclass[11pt,a4paper]{article}
\usepackage[hyperref]{acl2021}
\usepackage{times}
\usepackage{latexsym}
\usepackage{epsfig}
\usepackage{graphicx}
\usepackage{amsmath}
\usepackage{amssymb}
\usepackage{url}
\usepackage{booktabs}
\usepackage{subfigure}
\usepackage{bm}
\usepackage{color}
\usepackage{pifont}
\usepackage{amsfonts}
\usepackage{stmaryrd}
\usepackage{stfloats}
\usepackage{xspace}
\usepackage{threeparttable}
\usepackage{adjustbox}
\usepackage{algorithm,algorithmicx}
\usepackage{algpseudocode}
\usepackage{multirow}
\usepackage{soul}
\usepackage{ragged2e}
\usepackage{enumitem}

\usepackage{microtype}

\aclfinalcopy 
\def\aclpaperid{564} 

\newcommand{\paratitle}[1]{\vspace{0.5ex}\noindent\textbf{#1}}
\newcommand{\ie}{\emph{i.e.,}\xspace}
\newcommand{\eg}{\emph{e.g.,}\xspace}

\title{Parallel Attention Network with Sequence Matching for Video Grounding}

\author{
   Hao Zhang$^{1,2}$,~~Aixin Sun$^{1}$,~~Wei Jing$^{2,3}$\\
   \textbf{Liangli Zhen}$^{2}$,~~\textbf{Joey Tianyi Zhou}$^{2}$,~~\textbf{Rick Siow Mong Goh}$^{2}$ \\
   $^{1}$School of Computer Science and Engineering, Nanyang Technological University, Singapore\\
   $^{2}$Institute of High Performance Computing, A*STAR, Singapore\\
   $^{3}$Institute for Infocomm Research, A*STAR, Singapore\\
   \texttt{\{hao007@e.,axsun@\}ntu.edu.sg,}~~
   \texttt{21wjing@gmail.com,}\\
   \texttt{\{zhenll,zhouty,gohsm\}@ihpc.a-star.edu.sg}
}

\date{}

\begin{document}
\maketitle
\begin{abstract}
Given a video, video grounding aims to retrieve a temporal moment that semantically corresponds to a language query. In this work, we propose a \textbf{P}arallel \textbf{A}ttention \textbf{N}etwork with \textbf{Seq}uence matching (SeqPAN) to address the challenges in this task: multi-modal representation learning, and target moment boundary prediction. We design a self-guided parallel attention module to effectively capture self-modal contexts and cross-modal attentive information between video and text. Inspired by sequence labeling tasks in natural language processing, we split the ground truth moment into begin, inside, and end regions. We then propose a sequence matching strategy to guide start/end boundary predictions using region labels. Experimental results on three datasets show that SeqPAN is superior to state-of-the-art methods. Furthermore, the effectiveness of the self-guided parallel attention module and the sequence matching module is verified.\footnote{https://github.com/IsaacChanghau/SeqPAN}
\end{abstract}

\section{Introduction}\label{sec:intro}
\begin{figure}[t]
    \centering
    \includegraphics[width=0.47\textwidth]{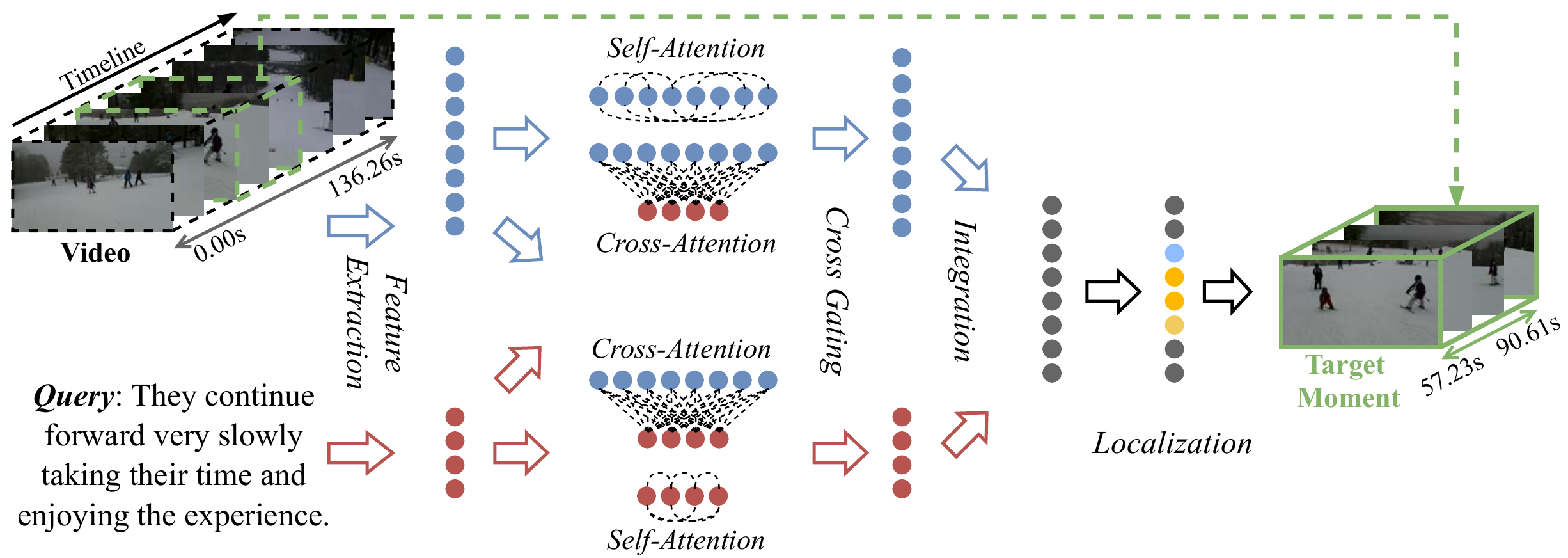}
    \caption{\small The overview of our procedures for video grounding, with an example of retrieving the temporal moment from an untrimmed video by a given language query.}
	\label{fig:example}
\end{figure}

Video grounding is a fundamental and challenging problem in vision-language understanding research area~\cite{hu2019looking,yu2019activityqa,zhu2020actbert}. It aims to retrieve a temporal video moment that semantically corresponds to a given language query, as shown in Figure~\ref{fig:example}. This task requires techniques from both computer vision~\cite{tran2015learning,shou2016temporal,feichtenhofer2019slowfast}, natural language processing~\cite{yu2018fast,yang2019xlNet}, and more importantly, the cross-modal interactions between the two. Many existing solutions~\cite{chen2018temporally,Liu2018TemporalMN,Xu2019MultilevelLA} tackle video grounding problem with \textit{proposal-based} approach. This approach generates proposals with pre-set sliding windows or anchors, computes the similarity between the query and each proposal. The proposal with highest score is selected as the answer. These methods are sensitive to the quality of proposals and are inefficient because all proposal-query pairs are compared. Recently, several one-stage \textit{proposal-free} solutions~\cite{chen2019localizing,lu2019debug,mun2020local} are proposed to directly predict start/end boundaries of target moments, through modeling video-text interactions. Our solution, SeqPAN, is a proposal-free method; hence our key focuses are \textit{video-text interaction modeling} and \textit{moment boundary prediction}.

\noindent\textbf{Video-text interaction modeling.} In order to model video-text interaction, various attention-based methods have been proposed~\cite{Gao2017TALLTA,yuan2019semantic,mun2020local}. In particular, transformer block~\cite{vaswani2017attention} is widely used in vision-language tasks and proved to be effective for multimodal learning~\cite{tan2019lxmert,lu2019vilbert,su2020vlbert,li2020hero}. In video grounding task, fine-grain scale unimodal representations are important to achieve good localization performance. However, existing solutions do not refine unimodal representations of video and text when doing cross-modal reasoning, and thus limit the performance.

To better capture informative features for multi-modalities, we encode both self-attentive contexts and cross-modal interactions from video and query. That is, instead of solely relying on sophisticated cross-modal learning as in most existing studies, we learn both intra- and inter-modal representations simultaneously, with improved attention modules.

\noindent\textbf{Moment boundary prediction.} In terms of the length, target moment is usually a very small portion of the video, making positive (frames in target moment) and negative (frames not in target moment) samples imbalanced. Further, we aim to predict the exact start/end boundaries (\ie two video frames\footnote{The ``frame'' is a general description, which can refer to a frame in a video sequence or a unit in the corresponding video feature representation.}) of the target moment. If we view from the space of video frames, sparsity is a major concern, \eg catching two frames among thousands. Recent studies attempt to address this issue by auxiliary objectives, \eg to discriminate whether each frame is foreground (positive) or background (negative)~\cite{Yuan2019ToFW,mun2020local}, or to regress distances of each frame within target moment to ground truth boundaries~\cite{lu2019debug,zeng2020dense}. However, the ``sequence'' nature of frames or videos is not considered.

\begin{figure}[t]
    \centering
    \includegraphics[width=0.45\textwidth]{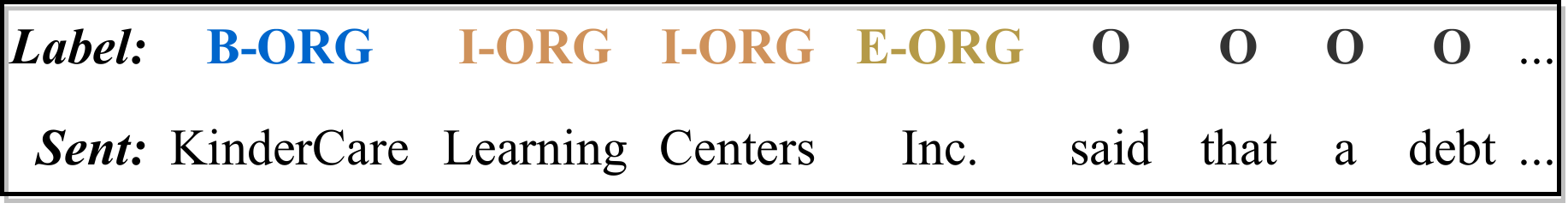}
    \caption{\small An example of the annotations in NER, where ``ORG'' is for ``Organization'', ``B'', ``I'' and ``E'' denote the begin, inside and end of the organization entity, respectively.}
	\label{fig:ner_example}
\end{figure}

We emphasize the ``sequence'' nature of video frames and adopt the concept of sequence labeling in NLP to video grounding.  We use named entity recognition (NER)~\cite{lample2016neural,ma2016end} as an example sequence labeling task for illustration in Figure~\ref{fig:ner_example}. Video grounding is to retrieve a sequence of frames with start/end boundaries of target moment from video. This is analogous to extract a multi-word named entity from a sentence. The main difference is that, words are discrete, so word annotations (\ie B, I, E, and O tags) in sentence are discrete. In contrast, video is continuous and the changes between consecutive frames are smooth. Hence, it is difficult (and also not necessary) to precisely annotate each frame. We relax the annotations on video sequence by specifying video regions, instead of frames. With respect to the target moment, we label B, I, E and O (BIEO) regions on video (see Figure~\ref{fig:architecture}) and introduce label embeddings to model these regions.

\noindent\textbf{Our contributions.} In this research, we propose a \textbf{P}arallel \textbf{A}ttention \textbf{N}etwork with \textbf{Seq}uence matching (SeqPAN) for video grounding task. 
We first design a \textit{self-guided parallel attention (SGPA)} module to capture both self- and cross-modal attentive information for each modality simultaneously. In SGPA module, a cross-gating strategy with self-guided head is further used to fuse self- and cross-modal representations. We then propose a \textit{sequence matching (sq-match)} strategy, to identify BIEO regions in video. The label embeddings are incorporated to represent label of frames in each region for region recognition. The sq-match guides SeqPAN to search for boundaries of target moment within constrained regions, leading to more precise localization results. Experimental results on three benchmarks demonstrate that both SGPA and sq-match consistently improve the performance; and SeqPAN surpasses the state-of-the-art methods.

\section{Related Work}\label{sec:related}
Existing solutions to video grounding are roughly categorized into proposal-based and proposal-free frameworks. In proposal-based framework, common structures include ranking and anchor-based methods. \textit{Ranking-based} methods~\cite{Liu2018CML,hendricks2017localizing,hendricks2018localizingM,chen2019semantic,ge2019mac,zhang2019exploiting} solve this task with two-stage propose-and-rank pipeline, which first generates proposals and then uses multimodal matching to retrieve most similar proposal for a query. \textit{Anchor-based} methods~\cite{chen2018temporally,yuan2019semantic,zhu2019cross,Wang2020TemporallyGL} sequentially assign each frame with multiscale temporal anchors and select the anchor with highest confidence as the result. However, these methods are sensitive to the proposal quality; and comparison of all proposal-query pairs is computational expensive and inefficient.

Proposal-free framework includes regression and span-based methods. \textit{Regression-based} methods~\cite{Yuan2019ToFW,lu2019debug,chen2020rethinking,chen2020learning} tackle video grounding by learning cross-modal interactions between video and query, and directly regressing temporal time of target moments. 
\textit{Span-based} methods~\cite{ghosh2019excl,rodriguez2020proposal,zhang2020vslnet,lei2020tvr,zhang2021qa4nlvl} address video grounding by borrowing the concept of extractive question answering~\cite{Seo2017BidirectionalAF,huang2018fusionnet}, and to predict the start and end boundaries of target moment directly. 

In addition, there are several works~\cite{he2019Readwa,Wang2019LanguageDrivenTA,cao2020strong,hahn2020tripping,wu2020reinforcement,Wu2020TreeStructuredPB} that formulate this task as a sequential decision-making problem and adopt reinforcement learning to observe candidate moments conditioned on queries. Other methods, \eg weakly supervised learning methods~\cite{mithun2019weakly,Lin2020WeaklySupervisedVM,wu2020reinforcement}, 2D map model of temporal relations between video moments~\cite{zhang2020learning}, ensemble of top-down and bottom-up methods~\cite{wang2020dual}, joint learning video-level matching and moment-level localization~\cite{shao2018find}, have also been explored. Some works~\cite{shao2018find,cao2020strong,liu2020jointly,wang2020dual} use either additional resources/features or different evaluation metrics, so their results are not directly comparable with many others, including ours.

\section{Proposed Method}

\begin{figure*}[t]
    \centering
    \includegraphics[trim={0cm 0.12cm 0cm 0cm},clip,width=\textwidth]{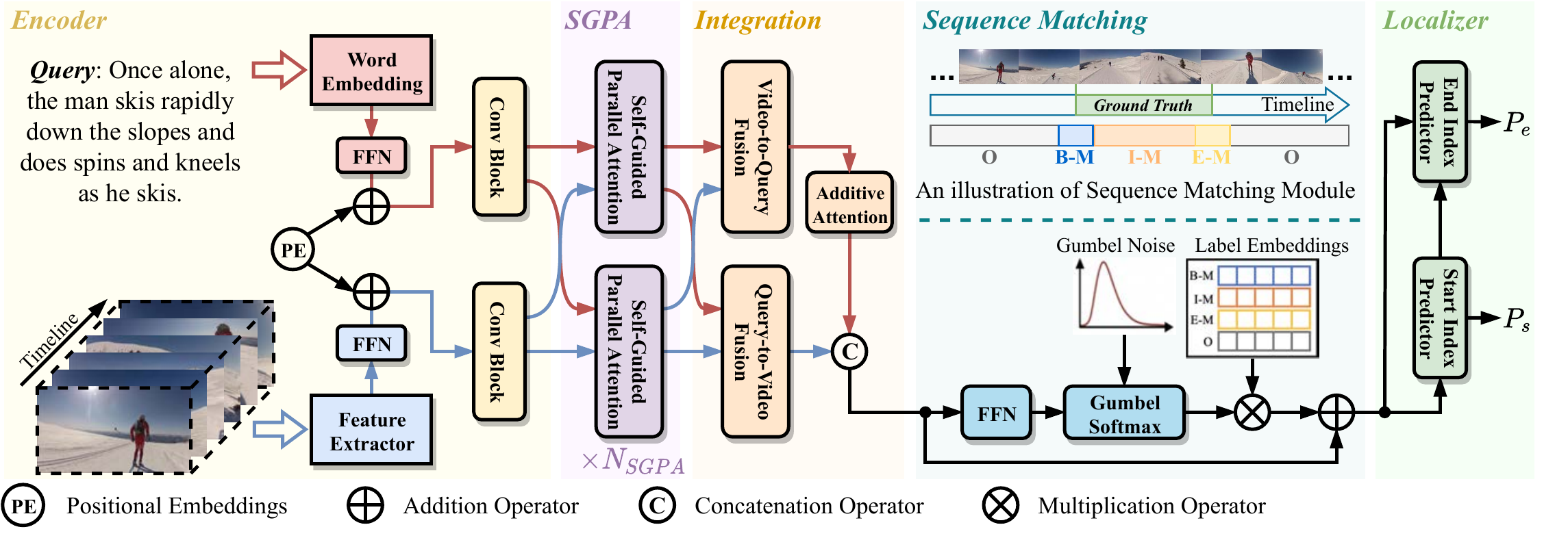}
    \caption{\small The architecture of the Parallel Attention Network with Sequence Matching (SeqPAN) for video grounding.}
	\label{fig:architecture}
\end{figure*}

Let $V=[f_t]_{t=0}^{T-1}$ be an untrimmed video with $T$ frames; $Q=[q_j]_{j=0}^{M-1}$ be a language query with $M$ words; $t^s$ and $t^e$ denote start and end time point of ground-truth temporal moment. We define and tackle video grounding task in feature spaces. Specifically, we split the given video $V$ into $N$ clip units, and use pre-trained feature extractor to encode them into visual features $\bm{V}=[\bm{v}_i]_{i=0}^{N-1}\in\mathbb{R}^{d_v\times N}$, where $d_v$ is visual feature dimension. Then the $t^{s(e)}$ are mapped to the corresponding indices $i^{s(e)}$ in the feature sequence, where $0\leq i^s \leq i^e \leq N-1$. For the query $Q$, we encode words with pre-trained word embeddings as $\bm{Q}=[\bm{w}_j]_{j=0}^{M-1}\in\mathbb{R}^{d_w\times M}$, where $d_w$ is word dimension. Given the pair of $(\bm{V},\bm{Q})$ as input, video grounding aims to localize a temporal moment starting at $i^s$ and ending at $i^e$.

\subsection{The SeqPAN Model}
The overall architecture of the proposed SeqPAN model is shown in Figure~\ref{fig:architecture}. Next, we present each module of SeqPAN in detail. 

\subsubsection{Encoder Module}\label{encoder_module}
Given visual features $\bm{V}\in\mathbb{R}^{d_v\times N}$ of the video and word embeddings $\bm{Q}\in\mathbb{R}^{d_w\times M}$ of the language query, we map them into the same dimension $d$ with two FFNs\footnote{We denote the single-layer feed-forward network as FFN ($\mathtt{FFN}(\bm{X})=\bm{W}\cdot\bm{X}+\bm{b}$) in this work.}, respectively. The encoder module mainly encodes the individual modality separately. As position encoding offers a flexible way to embed a sequence, when the sequence order matters, we first incorporate a position embedding to every input of both video and query sequences. Then, we adopt stacked 1D convolutional block to learn representations by carrying knowledge from neighbor tokens. The encoded representations are written as:
\begin{equation}
\begin{aligned}
    \bm{V'} & = \mathtt{ConvBlock}(\mathtt{FFN_v}(\bm{V})+\bm{E}_p) \\
    \bm{Q'} & = \mathtt{ConvBlock}(\mathtt{FFN_q}(\bm{Q})+\bm{E}_p)
\end{aligned}
\label{eq:encoder}
\end{equation}
where $\bm{V'}\in\mathbb{R}^{d\times N}$ and $\bm{Q'}\in\mathbb{R}^{d\times M}$; $\bm{E}_p$ denotes the positional embeddings. Both position embeddings and convolutional block are shared by the video and text features.

\subsubsection{Self-Guided Parallel Attention Module}\label{sgpa_module}
\begin{figure}[t]
    \centering
    \includegraphics[width=0.47\textwidth]{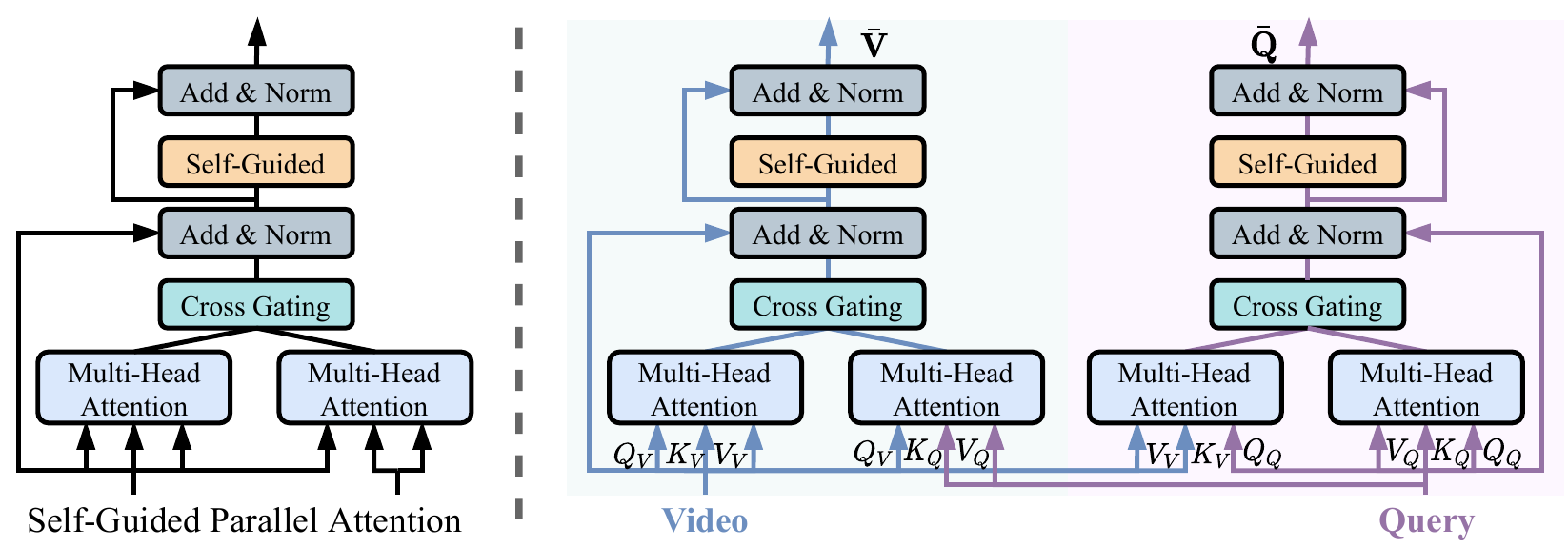}
    \caption{\small Self-Guided Parallel Attention (SGPA). Left: the structure of SGPA; Right: the parallel streams of encoding visual and textual inputs.}
	\label{fig:sgpa}
\end{figure}

A self-guided parallel attention (SGPA) module (see Figure~\ref{fig:sgpa}) is proposed to improve multimodal representation learning.
Compared with standard transformer (TRM) encoder, SGPA uses two parallel multi-head attention blocks to learn both \textit{uni-modal} and \textit{cross-modal} representations simultaneously, and merge them with a cross-gating strategy\footnote{A detailed comparison of SGPA and standard TRMs is summarized in Appendix.}. Taking video modality as an example, the attention process is computed as:
\begin{equation}
\begin{aligned}
    \bm{\hat{V}}_{\text{S}}= V_V\cdot\sigma_{s}\Big(\frac{Q^{\top}_V K_V}{\sqrt{d}}\Big)\\ \bm{\hat{V}}_{\text{C}}=V_Q\cdot\sigma_{s}\Big(\frac{Q^{\top}_V K_Q}{\sqrt{d}}\Big)
\end{aligned}
\label{eq:self_cross_attn}
\end{equation}
where $\sigma_{s}$ denotes Softmax function; $Q_V$, $K_V$ and $V_V$ are the linear projections of $\bm{V'}$; $Q_Q$, $K_Q$ and $V_Q$ are linear projections of $\bm{Q'}$; $\bm{\hat{V}}_{\text{S}}$ encodes the self-attentive contexts within video modality; and $\bm{\hat{V}}_{\text{C}}$ integrates information from query modality according to cross-modal attentive relations. 
The self- and cross-modal representations are then merged together by a cross-gating strategy:
\begin{equation}
    \bm{\hat{V}} = \sigma\big(\mathtt{FFN}(\bm{\hat{V}}_{\text{C}})\big)\odot\bm{\hat{V}}_{\text{S}} + \sigma\big(\mathtt{FFN}(\bm{\hat{V}}_{\text{S}})\big)\odot\bm{\hat{V}}_{\text{C}}
\label{eq:cross_gate}
\end{equation}
where $\sigma$ denotes Sigmoid function and $\odot$ represents Hadamard product. The cross-gating explicitly interacts features obtained from the self- and cross-attention encoders to ensure both are fully utilized, instead of relying on only one of them. Finally, we employ a self-guided head to implicitly emphasize the informative representations by measuring the confidence of each element in $\bm{\hat{V}}$ as:
\begin{equation}
    \bm{\bar{V}} = \sigma\big(\mathtt{FFN}_{\sigma}(\bm{\hat{V}})\big)\odot\mathtt{FFN}(\bm{\hat{V}})
\label{eq:self_guided}
\end{equation}

The refined representations $\bm{\bar{Q}}$ for the query modality are obtained in a similar manner (\eg swapping visual and query features). 

\subsubsection{Video-Query Integration Module}\label{vqi_module}
This module further enhances the cross-modal interactions between visual and textual features. It utilizes context-query attention (CQA) strategy~\cite{yu2018fast} and aggregates text information for each visual element\footnote{We provide a detailed computation process in appendix.} (see Figure~\ref{fig:architecture}). Given $\bm{\bar{V}}$ and $\bm{\bar{Q}}$, CQA first computes the similarities, $\mathcal{S}\in\mathbb{R}^{N\times M}$, between each pair of $\bm{\bar{V}}$ and $\bm{\bar{Q}}$ features. Then two attention weights are derived by $\mathcal{A}_{\text{VQ}}=\mathcal{S}_r\cdot\bm{\bar{Q}}^{\top}$ and $\mathcal{A}_{\text{QV}}=\mathcal{S}_r\cdot\mathcal{S}^{\top}_c\cdot\bm{\bar{V}}^{\top}$, where $\mathcal{S}_{r}$/$\mathcal{S}_{c}$ are row-/column-wise normalization of $\mathcal{S}$ by Softmax. The query-aware video representations $\bm{V}^{Q}$ is computed by:
\begin{equation}
    \bm{V}^{Q}=\mathtt{FFN}\big([\bm{\bar{V}};\mathcal{A}_{\text{VQ}}^{\top};\bm{\bar{V}}\odot\mathcal{A}_{\text{VQ}}^{\top};\bm{\bar{V}}\odot\mathcal{A}_{\text{QV}}^{\top}]\big)
\label{eq:tri_attn}
\end{equation}
where $\bm{V}^{Q}\in\mathbb{R}^{d\times N}$. Similarly, video-aware query representations $\bm{Q}^{V}\in\mathbb{R}^{d\times M}$ can be derived by swapping visual and textual inputs in CQA module. Then we encode $\bm{Q}^{V}$ into sentence representation $\bm{q}$ with additive attention~\cite{Bahdanau2015NeuralMT} and concatenate it with each element of $\bm{V}^{Q}$ as $\bm{H}=[\bm{h}_{1},\dots,\bm{h}_{n}]$, where $\bm{h}_{i}=[\bm{v}_{i}^{Q};\bm{q}]$. Finally, the query-attended visual representation is computed as $\bm{\bar{H}}=\mathtt{FFN}(\bm{H})\in\mathbb{R}^{d\times N}$.

\subsubsection{Sequence Matching Module}\label{seq_match_module}
As illustrated in Figure~\ref{fig:architecture}, we considers the frames within ground truth moment and several neighboring frames as foreground, while the rest as background. Then, we split the foreground into \textbf{B}egin, \textbf{I}nside, and \textbf{E}nd regions. For simplicity, we assign each region a label, \ie ``B-M'' for begin, ``I-M'' for inside, ``E-M'' for end region, and ``O'' for background. B-M/E-M explicitly indicate potential positions of the start/end boundaries. We also specify orthogonal label embeddings $\bm{E}_{\text{lab}}\in\mathbb{R}^{d\times 4}$ to represent those labels, and to infuse label information into visual features after region label predictions.

Note our approach is different from~\citet{lin2018bsn} on temporal action proposal generation task, where the target proposal is split into start, centre, and end regions. The probability of a frame belonging to each of three regions is predicted separately in a regression manner, leading to three separate probability sequences, one for each region. The maximum probabilities in the sequences are used to guide proposal generations. In contrast, we formulate matching process as a multi-class classification problem and predict a concrete region label for each frame, \ie same as a sequence labeling task in NLP. Label embeddings are then assigned to the frames based on the labels of the predicted region.

A straightforward solution to predict the confidence of an element belonging to each region is multi-class classifier:
\begin{equation}
    \bm{H}_{\text{seq}}=\mathtt{FFN}_{\text{seq}}(\bm{\bar{H}}),\;\;
    \bm{S}_{\text{seq}} = \sigma_s(\bm{H}_{\text{seq}})\in\mathbb{R}^{4\times N}
\label{eq:seq_softmax}
\end{equation}
where $\bm{S}_{\text{seq}}$ encodes the probabilities of each visual element in different regions. Then label index with highest probability from $\bm{S}_{\text{seq}}$ is selected to represent the predicted label for each visual element:
\begin{equation}
    \bm{L}_{\text{lab}} = [\arg\max(\bm{S}^{j}_{\text{seq}})]_{j=0}^{N-1}\in\mathbb{R}^{N}
\label{eq:arg_max}
\end{equation}
However, a major issue here is that Eq.~\ref{eq:arg_max} needs to sample from a discrete probability distribution, which makes the back-propagation of gradients through $\bm{S}_{\text{seq}}$ in Eq.~\ref{eq:seq_softmax} infeasible for optimizer. To make back-propagation possible, we adopt the Gumbel-Max~\cite{gumbel1954statistical,maddison2014asampling} trick to re-formulate Eq.~\ref{eq:arg_max} as:
\begin{equation}
    \bm{\hat{L}}_{\text{lab}}=\big[\mathtt{Onehot}(\arg\max(\bm{H}^{j}_{\text{seq}}+\bm{g}))\big]_{j=0}^{N-1}
\label{eq:gumbel_max}
\end{equation}
where $\bm{\hat{L}}_{\text{lab}}\in\mathbb{R}^{4\times N}$. Then, we utilize the Gumbel-Softmax~\cite{jang2017categorical,maddison2017the} to relax the $\arg\max$ so as to make Eq.~\ref{eq:gumbel_max} being differentiable\footnote{More details about Gumbel Tricks are in Appendix.}. Formally, we use Eq.~\ref{eq:gumbel_softmax} to approximate Eq.~\ref{eq:gumbel_max} as:
\begin{equation}
    \bm{\bar{L}}_{\text{lab}} = \sigma_s\big((\bm{H}_{\text{seq}}+\bm{g})/\tau\big)\in\mathbb{R}^{4\times N}
\label{eq:gumbel_softmax}
\end{equation}
where $\tau$ is annealing temperature. As $\tau\to 0^{+}$, $\bm{\bar{L}}_{\text{lab}}\approx\bm{\hat{L}}_{\text{lab}}$, while $\tau\to\infty$, each element in $\bm{\bar{L}}_{\text{lab}}$ will be the same and the approximated distribution will be smooth. Note we use Eq.~\ref{eq:gumbel_max} during forward pass while Eq.~\ref{eq:gumbel_softmax} for backward pass to allow gradient back-propagation.
As the result, the embedding lookup process is differentiable and the label-attended visual representations is derived as:
\begin{equation}
    \bm{\widetilde{H}} = \bm{E}_{\text{lab}}\cdot\bm{\hat{L}}_{\text{lab}} + \bm{\bar{H}}
\label{eq:label_attend_H}
\end{equation}
The training objective is defined as:
\begin{equation}
    \mathcal{L}_{\text{seq}} = f_{\text{XE}}(\bm{\bar{L}}_{\text{lab}}, \bm{Y}_{\text{lab}}) + \Vert\bm{E}_{\text{lab}}^{\top}\bm{E}_{\text{lab}}\odot(\bm{1}-\bm{I})\Vert_{\text{F}}^{2}
\label{eq:seq_loss}
\end{equation}
where $\bm{Y}_{\text{lab}}$ denotes the ground truth sequence labels, $\bm{1}$ is the matrix with all elements being $1$ and $\bm{I}$ is the identity matrix. The second term in Eq.~\ref{eq:seq_loss} is the orthogonal regularization~\cite{brock2019large}, which ensures $\bm{E}_{\text{lab}}$ to keep the orthogonality.

\subsubsection{Localization Module}\label{localize_module}
Finally, we present a conditioned localizer to predict the start and end boundaries of the target moment. The localizer consists of two stacked transformer blocks and two FFNs. The scores of start and end boundaries are calculated as:
\begin{equation}
\begin{aligned}
    \bm{H}_{\text{s}} = \mathtt{TRM}_{s}(\bm{\widetilde{H}}), &\; \bm{S}_s = \bm{W}_s[\bm{H}_{\text{s}};\bm{\widetilde{H}}]+\bm{b}_s
    \\
    \bm{H}_{\text{e}} = \mathtt{TRM}_{e}(\bm{H}_{\text{s}}), &\; \bm{S}_e = \bm{W}_e[\bm{H}_{\text{e}};\bm{\widetilde{H}}]+\bm{b}_e
\end{aligned}
\label{eq:span_predictor}
\end{equation}
where $\bm{S}_{s/e}\in\mathbb{R}^{N}$. $\bm{W}_{s/e}$ and $\bm{b}_{s/e}$ are the weight and bias of start/end FFNs, respectively. Note the representations of end boundary ($\bm{H}_{\text{e}}$) are conditioned on that of start boundary ($\bm{H}_{\text{s}}$) to ensure the predicted end boundary is always after start boundary. Then, the probability distributions of start/end boundaries are computed by $\bm{P}_{s/e} = \mathtt{Softmax}(\bm{S}_{s/e})\in\mathbb{R}^{N}$. The training objective is:
\begin{equation}
    \mathcal{L}_{\text{loc}} = \frac{1}{2}\times\big[f_{\text{XE}}(\bm{P}_s, \bm{Y}_s) + f_{\text{XE}}(\bm{P}_e, \bm{Y}_e)\big]
\label{eq:loc_loss}
\end{equation}
where $f_{\text{XE}}$ is cross-entropy function, $\bm{Y}_{s/e}$ is one-hot labels for start/end ($i^s$/$i^e$) boundaries.

\subsection{Training and Inference}\label{train_infer}
The overall training loss of SeqPAN is: $\mathcal{L}=\mathcal{L}_{\text{loc}}+\mathcal{L}_{\text{seq}}$, to be minimized during the training process. During inference, the predicted start and end boundaries of a given video-query pair are generated by maximizing the joint probability as:
\begin{equation}
\begin{aligned}
    (\hat{i}^s,\hat{i}^e) = & \arg\max_{\hat{a}^s,\hat{a}^e} \bm{P}_s(\hat{a}^s)\times\bm{P}_e(\hat{a}^e) \\
    \text{s.t.: } & 0\leq\hat{i}^s\leq \hat{i}^e\leq N - 1
\end{aligned}
\label{eq:infer}
\end{equation}
where $\hat{i}^s$ and $\hat{i}^e$ are the best start and end boundaries of predicted moment for the given video-query pair. Let $\mathcal{T}$ be the duration of given video, the predicted start/end time are computed by $\hat{t}^{s(e)} = \hat{i}^{s(e)}/(N-1)\times\mathcal{T}$. With the predicted $(\hat{t}^s, \hat{t}^e)$ and ground truth $(t^s, t^e)$ time intervals, the measure, temporal intersection over union (IoU), is computed as:
\begin{equation}
    \text{IoU} = \max\Big(0, \frac{t_{\text{min}}^e-t_{\text{max}}^s}{t_{\text{max}}^e-t_{\text{min}}^s}\Big)\in [0, 1]
\label{eq:iou}
\end{equation}
where $t_{\min/\max}^{s(e)}=\min/\max(\hat{t}^{s(e)}, t^{s(e)})$.

\section{Experiments}

\subsection{Experimental Setting}
\noindent\textbf{Datasets.} We conduct the experiments on three benchmark datasets: Charades-STA~\cite{Gao2017TALLTA}, ActivityNet Captions~\cite{krishna2017dense} and TACoS~\cite{regneri2013grounding}. \textbf{Charades-STA}, collected by \citet{Gao2017TALLTA} from Charades~\cite{sigurdsson2016hollywood} dataset, contains $16,128$ annotations (\ie moment-query pairs), where $12,408$ and $3,720$ annotations are for train and test. \textbf{ActivityNet Captions} (\textbf{ANetCaps}) contains $20$K videos taken from ActivityNet~\cite{heilbron2015activitynet} dataset. We follow the setup in~\cite{chen2020rethinking,lu2019debug,Wu2020TreeStructuredPB,Yuan2019ToFW,zhang2020vslnet} with $37,421$ and $17,505$ annotations for train and test. \textbf{TACoS} contains $127$ cooking activities videos from~\citet{Rohrbach2012SDA}. We follow~\citet{Gao2017TALLTA} with $10,146$, $4,589$, and $4,083$ annotations are used for train, validation, and test, respectively.

\noindent\textbf{Evaluation Metric.} (i) ``$\text{R@}n, \text{IoU=}\mu$'', which denotes the percentage of test samples that have at least one result whose IoU with ground-truth is larger than $\mu$ in top-$n$ predictions; (ii) ``mIoU'', which denotes the average IoU over all test samples. We set $n=1$ and $\mu\in\{0.3, 0.5, 0.7\}$.

\noindent\textbf{Implementation Details.} We follow~\cite{ghosh2019excl,mun2020local,rodriguez2020proposal,zhang2020vslnet} and use 3D ConvNet pre-trained on Kinetics dataset~\cite{Carreira2017QuoVA} to extract RGB visual features from videos; then we downsample the feature sequence to a fixed length. The query words are lowercased and initialized with GloVe~\cite{pennington2014glove} embedding. We set hidden dimension $d$ to $128$; SGPA blocks to $2$; annealing temperature to $0.3$; and heads in multi-head attention to $8$; Adam~\cite{Kingma2015AdamAM} optimizer with batch size of $16$ and learning rate of $0.0001$ is used for training.

More details of dataset statistics and the hyper-parameter settings are summarized in Appendix.

\begin{table}[t]
    \small
    \centering
	\setlength{\tabcolsep}{7.0 pt}
	\begin{tabular}{l | c c c | c}
		\specialrule{.1em}{.05em}{.05em}
		\multirow{2}{*}{Methods} & \multicolumn{3}{c |}{$\text{R@}1, \text{IoU}=\mu$} & \multirow{2}{*}{mIoU} \\
        & $\mu=0.3$ & $\mu=0.5$ & $\mu=0.7$ & \\
        \hline
        DEBUG   & 54.95 & 37.39 & 17.69 & 36.34 \\
        ExCL    & 61.50 & 44.10 & 22.40 & -     \\
        MAN     & -     & 46.53 & 22.72 & -     \\
        SCDM    & -     & 54.44 & 33.43 & -     \\
        CBP     & -     & 36.80 & 18.87 & -     \\
        GDP     & 54.54 & 39.47 & 18.49 & -     \\
        2D-TAN  & -     & 39.81 & 23.31 & -     \\
        TSP-PRL & -     & 45.30 & 24.73 & 40.93 \\
        TMLGA   & 67.53 & 52.02 & 33.74 & -     \\
        VSLNet  & 70.46 & 54.19 & 35.22 & \textit{50.02} \\
        DRN     & -     & 53.09 & 31.75 & -     \\
        LGI     & \textit{72.96} & \textit{59.46} & \textit{35.48} & -     \\
        \hline
        SeqPAN  & \textbf{73.84} & \textbf{60.86} & \textbf{41.34} & \textbf{53.92} \\
        \specialrule{.1em}{.05em}{.05em}
	\end{tabular}
	\caption{\small Comparison with SOTA methods on Charades-STA.}
	\label{tab:sota_charades}
\end{table}

\begin{table}[t]
    \small
	\centering
	\setlength{\tabcolsep}{7.0 pt}
	\begin{tabular}{l | c c c | c}
		\specialrule{.1em}{.05em}{.05em}
		\multirow{2}{*}{Methods} & \multicolumn{3}{c|}{$\text{R@}1, \text{IoU}=\mu$} & \multirow{2}{*}{mIoU} \\
        & $\mu=0.3$ & $\mu=0.5$ & $\mu=0.7$ & \\
        \hline
        DEBUG   & 55.91 & 39.72 & -     & 39.51 \\
        ExCL    & \textit{63.00} & 43.60 & 24.10 & -     \\
        SCDM    & 54.80 & 36.75 & 19.86 & -     \\
        CBP     & 54.30 & 35.76 & 17.80 & -     \\
        GDP     & 56.17 & 39.27 & -     & 39.80 \\
        2D-TAN  & 59.45 & 44.51 & \textit{27.38} & -     \\
        TSP-PRL & 56.08 & 38.76 & -     & 39.21 \\
        TMLGA   & 51.28 & 33.04 & 19.26 & -     \\
        VSLNet  & \textbf{63.16} & 43.22 & 26.16 & \textit{43.19} \\
        DRN     & -     & \textit{45.45} & 24.36 & -     \\
        LGI     & 58.52 & 41.51 & 23.07 & -     \\
        \hline
        SeqPAN  & 61.65 & \textbf{45.50} & \textbf{28.37} & \textbf{45.11} \\
        \specialrule{.1em}{.05em}{.05em}
	\end{tabular}
	\caption{\small Comparison with SOTA methods on ANetCaps.}
	\label{tab:sota_activitynet}
\end{table}

\begin{table}[t]
    \small
	\centering
	\setlength{\tabcolsep}{7.0 pt}
	\begin{tabular}{l | c c c | c}
		\specialrule{.1em}{.05em}{.05em}
		\multirow{2}{*}{Methods} & \multicolumn{3}{c |}{$\text{R@}1, \text{IoU}=\mu$} & \multirow{2}{*}{mIoU} \\
        & $\mu=0.3$ & $\mu=0.5$ & $\mu=0.7$ & \\
        \hline
        TGN    & 21.77 & 18.90 & -     & -     \\
        ACL    & 24.17 & 20.01 & -     & -     \\
        DEBUG  & 23.45 & 11.72 & -     & 16.03 \\
        SCDM   & 26.11 & 21.17 & -     & -     \\
        CBP    & 27.31 & \textit{24.79} & 19.10 & 21.59 \\
        GDP    & 24.14 & 13.90 & -     & 16.18 \\
        TMLGA  & 24.54 & 21.65 & 16.46 & -     \\
        VSLNet & \textit{29.61} & 24.27 & \textit{20.03} & \textit{24.11} \\
        DRN    & -     & 23.17 & -     & -     \\
        \hline
        SeqPAN & \textbf{31.72} & \textbf{27.19} & \textbf{21.65} & \textbf{25.86} \\
        \hline\hline
        2D-TAN & 37.29 & 25.32 & -     & -     \\
        \hline
        SeqPAN & \textbf{48.64} & \textbf{39.64} & \textbf{28.07} & \textbf{37.17} \\
        \specialrule{.1em}{.05em}{.05em}
	\end{tabular}
	\caption{\small Comparison with SOTA methods on TACoS.}
	\label{tab:sota_tacos}
\end{table}

\begin{table}[t]
    \small
	\centering
	\begin{tabular}{l | c c c}
		\specialrule{.1em}{.05em}{.05em}
		\multirow{2}{*}{Methods} & \multicolumn{3}{c}{$\text{R@}1, \text{IoU}=\mu$} \\
        & $\mu=0.3$ & $\mu=0.5$ & $\mu=0.7$ \\
        \hline
        \multicolumn{4}{c}{Charades-STA} \\
        \hline
        Se-TRM & 68.84~\scriptsize{(0.46)} & 51.92~\scriptsize{(0.54)} & 34.58~\scriptsize{(0.18)} \\
        Co-TRM & 69.03~\scriptsize{(0.49)} & 52.34~\scriptsize{(0.50)} & 35.07~\scriptsize{(0.32)} \\
        SGPA   & 69.47~\scriptsize{(0.32)} & 54.63~\scriptsize{(0.43)} & 36.36~\scriptsize{(0.24)} \\
        \hline
        \multicolumn{4}{c}{ActivityNet Captions} \\
        \hline
        Se-TRM & 57.64~\scriptsize{(0.38)} & 40.76~\scriptsize{(0.35)} & 25.10~\scriptsize{(0.30)} \\
        Co-TRM & 57.39~\scriptsize{(0.29)} & 40.55~\scriptsize{(0.45)} & 24.85~\scriptsize{(0.47)} \\
        SGPA   & 58.40~\scriptsize{(0.31)} & 41.72~\scriptsize{(0.19)} & 26.07~\scriptsize{(0.16)} \\
        \specialrule{.1em}{.05em}{.05em}
	\end{tabular}
	\caption{\small SGPA vs. standard transformers on Charades-STA and ANetCaps. Se-TRM is the transformer block with single modality inputs, and Co-TRM~\cite{tan2019lxmert,lu2019vilbert,lei2020tvr,li2020hero} is with dual modality inputs. Scores in brackets are standard deviation.}
	\label{tab:transformer}
\end{table}

\subsection{Comparison with State-of-the-Arts}
We compare SeqPAN with the following state-of-the-arts. 1) \textit{Proposal-based} methods: TGN~\cite{chen2018temporally}, ACL~\cite{ge2019mac}, CBP~\cite{Wang2020TemporallyGL}, SCDM~\cite{yuan2019semantic}, MAN~\cite{zhang2019man}; 2) \textit{Proposal-free} methods: DEBUG~\cite{lu2019debug}, ExCL~\cite{ghosh2019excl}, VSLNet~\cite{zhang2020vslnet}, GDP~\cite{chen2020rethinking}, LGI~\cite{mun2020local}, TMLGA~\cite{rodriguez2020proposal}, DRN~\cite{zeng2020dense}; 3) \textit{Others}: TSP-PRL~\cite{Wu2020TreeStructuredPB} and 2D-TAN~\cite{zhang2020learning}. The best results are in \textbf{bold} and the second bests are in \textit{italic}. In all result tables, the scores of compared methods are reported in the corresponding works.

The results on the Charades-STA are summarized in Table~\ref{tab:sota_charades}. SeqPAN surpasses all baselines and achieves the highest scores over all metrics. Observe that the performance improvements of SeqPAN are more significant under more strict metrics. The results show that SeqPAN can produce more precise localization results. For instance, compared to LGI, SeqPAN achieves $5.86\%$ absolute improvement by ``R@1, IoU=0.7'', and $1.40\%$ by ``R@1, IoU=0.5''. Table~\ref{tab:sota_activitynet} reports the results on ANetCaps. SeqPAN is superior to baselines and achieves the best performance on ``R@1, IoU=0.7'' and mean IoU. As reported in Table~\ref{tab:sota_tacos}, similar observations hold on TACoS. Note 2D-TAN~\cite{zhang2020learning} pre-processes the TACoS dataset, making it is slightly different from the original one. We also conduct experiments on their version for a fair comparison. SeqPAN outperforms the baselines over all evaluation metrics on both versions.

\begin{table*}[t]
    \small
	\centering
	\setlength{\tabcolsep}{2.0 pt}
	\begin{tabular}{l | c c | c c c | c | c c c | c }
		\specialrule{.1em}{.05em}{.05em}
		\multirow{3}{*}{Method} & \multicolumn{2}{|c|}{\multirow{2}{*}{sq-match}} & \multicolumn{4}{c|}{Charades-STA} & \multicolumn{4}{c}{ActivityNet Captions} \\
		\cline{4-11}
		& & & \multicolumn{3}{c |}{$\text{R@}1, \text{IoU}=\mu$} & \multirow{2}{*}{mIoU} & \multicolumn{3}{c |}{$\text{R@}1, \text{IoU}=\mu$} & \multirow{2}{*}{mIoU} \\
		\cline{2-3}
        & \;\;$G$\; & $\bm{E}_{\text{lab}}$ & $\mu=0.3$ & $\mu=0.5$ & $\mu=0.7$ &  & $\mu=0.3$ & $\mu=0.5$ & $\mu=0.7$ &  \\
        \hline
        SeqPAN w/ fb-match & - & - & 70.27\scriptsize{(0.75)} & 56.96\scriptsize{(0.46)} & 38.95\scriptsize{(0.27)} & 51.84\scriptsize{(0.40)} & \textit{59.99}\scriptsize{(0.25)} & 43.71\scriptsize{(0.19)} & 26.72\scriptsize{(0.29)} & 43.23\scriptsize{(0.23)} \\
        \hline
        SeqPAN w/o sq-match & \ding{56} & \ding{56} & 69.62\scriptsize{(0.54)} & 55.29\scriptsize{(0.30)} & 36.71\scriptsize{(0.48)} & 51.13\scriptsize{(0.25)} & 59.03\scriptsize{(0.35)} & 42.65\scriptsize{(0.32)} & 26.29\scriptsize{(0.13)} & 42.51\scriptsize{(0.36)} \\
        SeqPAN w/ Gumbel & \ding{52} & \ding{56} & \textit{71.64}\scriptsize{(0.64)} & \textit{57.61}\scriptsize{(0.26)} & \textit{39.26}\scriptsize{(0.31)} & \textit{52.15}\scriptsize{(0.45)} & 59.74\scriptsize{(0.42)} & \textit{43.85}\scriptsize{(0.35)} & \textit{27.12}\scriptsize{(0.20)} & \textit{43.69}\scriptsize{(0.24)} \\
        SeqPAN & \ding{52} & \ding{52} & \textbf{72.70}\scriptsize{(0.51)} & \textbf{60.15}\scriptsize{(0.50)} & \textbf{41.02}\scriptsize{(0.36)} & \textbf{53.19}\scriptsize{(0.38)} & \textbf{61.12}\scriptsize{(0.39)} & \textbf{45.09}\scriptsize{(0.37)} & \textbf{27.97}\scriptsize{(0.27)} & \textbf{44.77}\scriptsize{(0.23)} \\
        \specialrule{.1em}{.05em}{.05em}
	\end{tabular}
	\caption{\small Ablation studies of sequence matching strategy in SeqPAN, where the values in bracket denote standard deviation.}
	\label{tab:seq_mat}
\end{table*}

\begin{figure}[t]
    \centering
	\subfigure[\small Charades-STA]
	{\label{fig:sgpa_block_charades}	\includegraphics[width=0.47\textwidth]{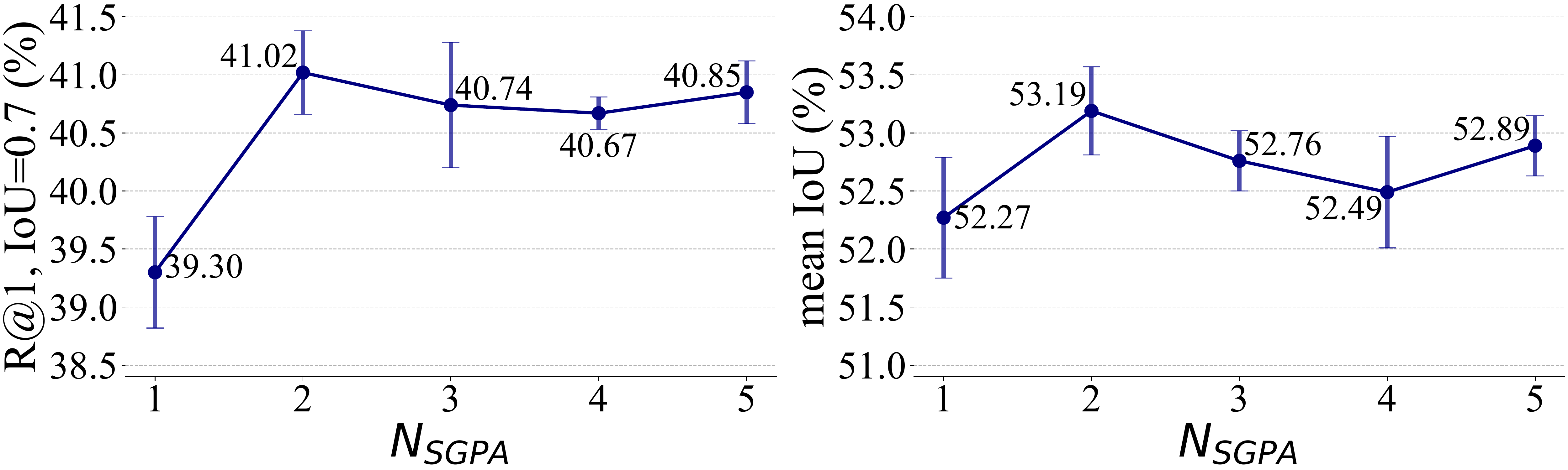}}
	\subfigure[\small ActivityNet Captions]
	{\label{fig:sgpa_block_activitynet}	\includegraphics[width=0.47\textwidth]{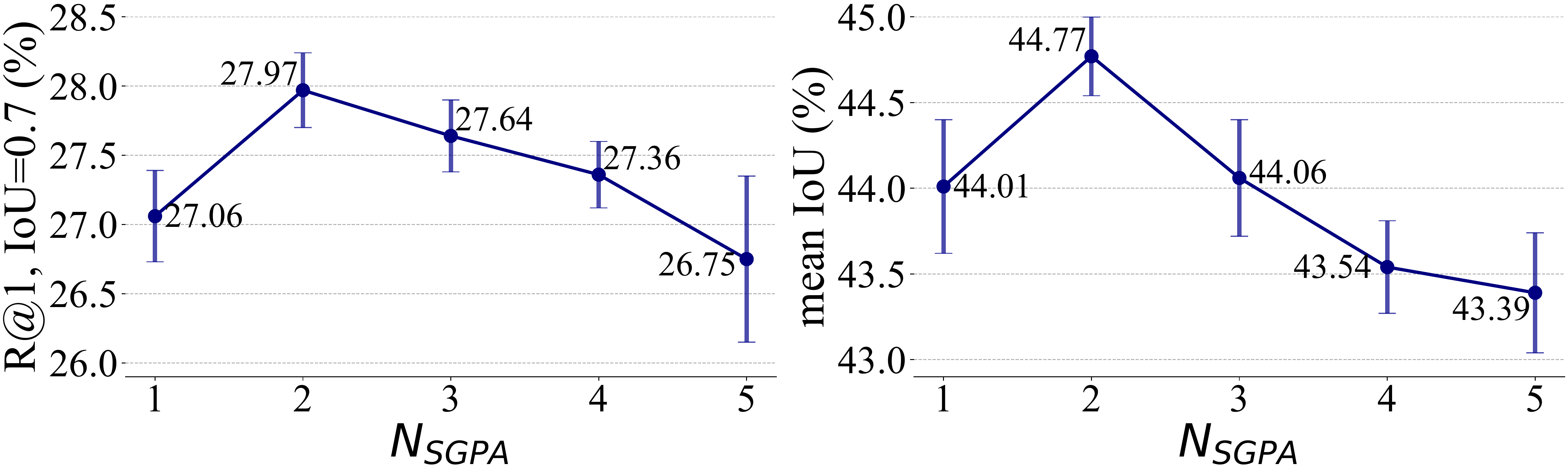}}
	\caption{\small The impact of SGPA block numbers ($N_{\text{SGPA}}$) on Charades-STA and ANetCaps.}
	\label{fig:sgpa_block}
\end{figure}

\begin{figure}[t]
    \centering
	\subfigure[\small Charades-STA]
	{\label{fig:tau_charades}	\includegraphics[width=0.47\textwidth]{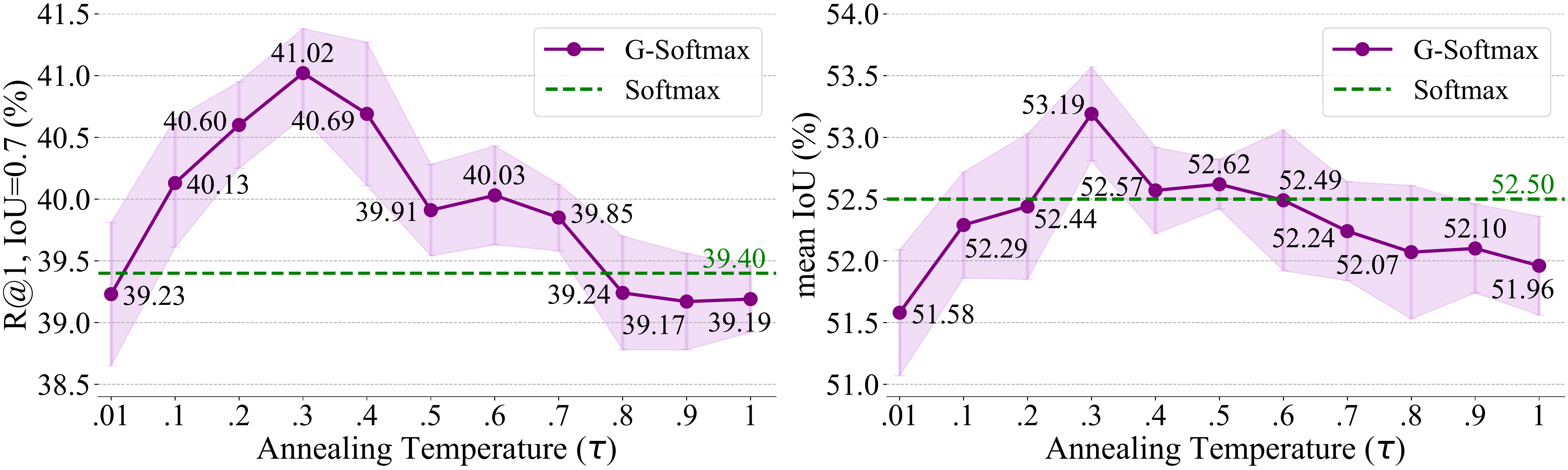}}
	\subfigure[\small ActivityNet Captions]
	{\label{fig:tau_activitynet}	\includegraphics[width=0.47\textwidth]{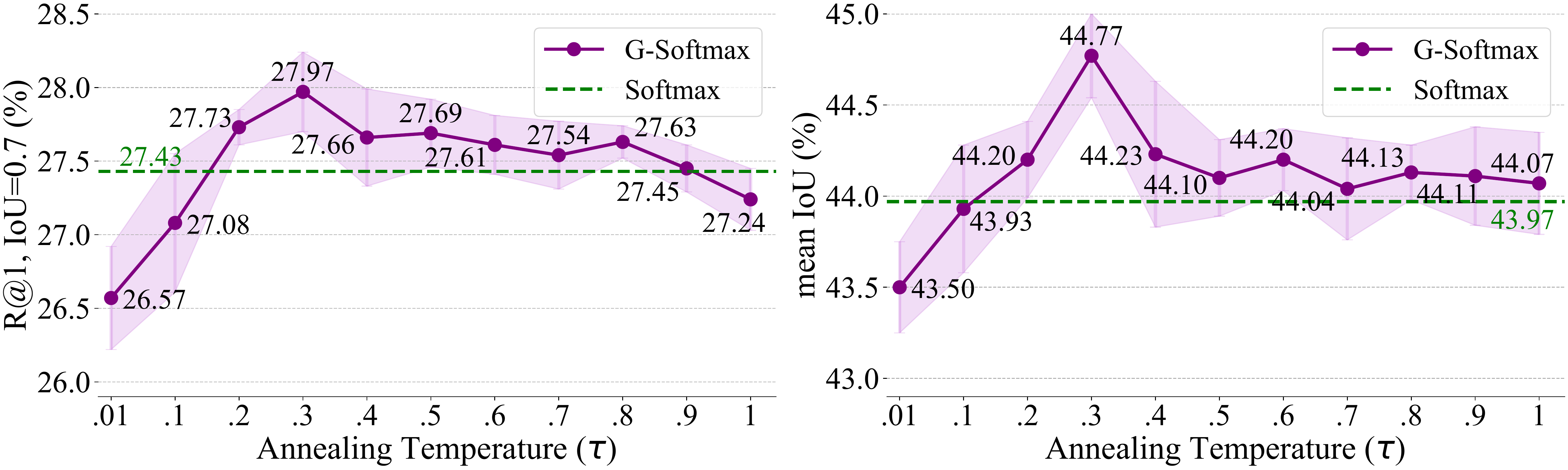}}
	\caption{\small The impact of annealing temperature $\tau$ in sequence matching on Charades-STA and ANetCaps.}
	\label{fig:annealing_temperature}
\end{figure}

\begin{figure}[t]
    \centering
	\subfigure[\small Charades-STA]
	{\label{fig:iou_count_charades}	\includegraphics[width=0.23\textwidth]{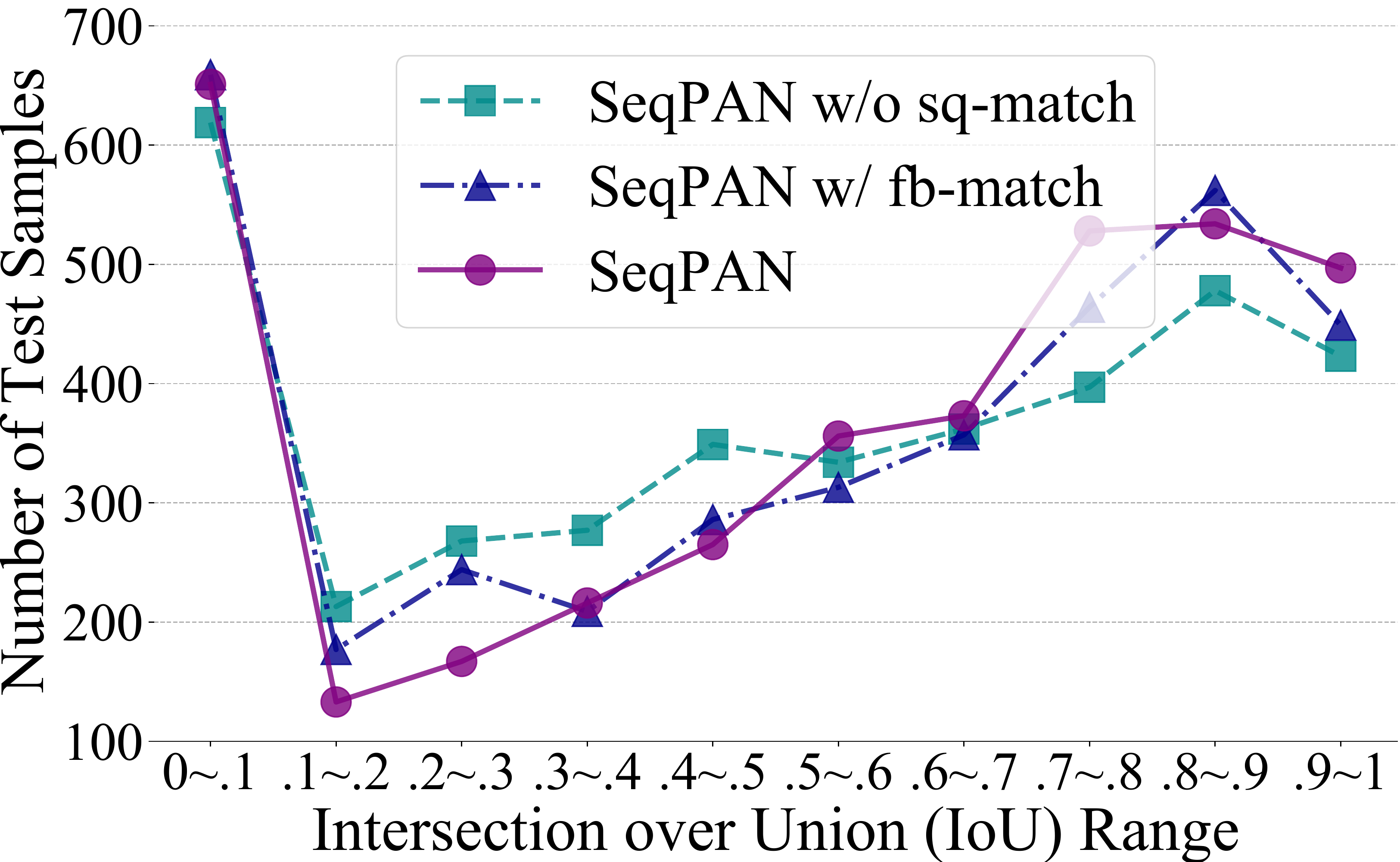}}
	\subfigure[\small ActivityNet Captions]
	{\label{fig:iou_count_activitynet}	\includegraphics[width=0.23\textwidth]{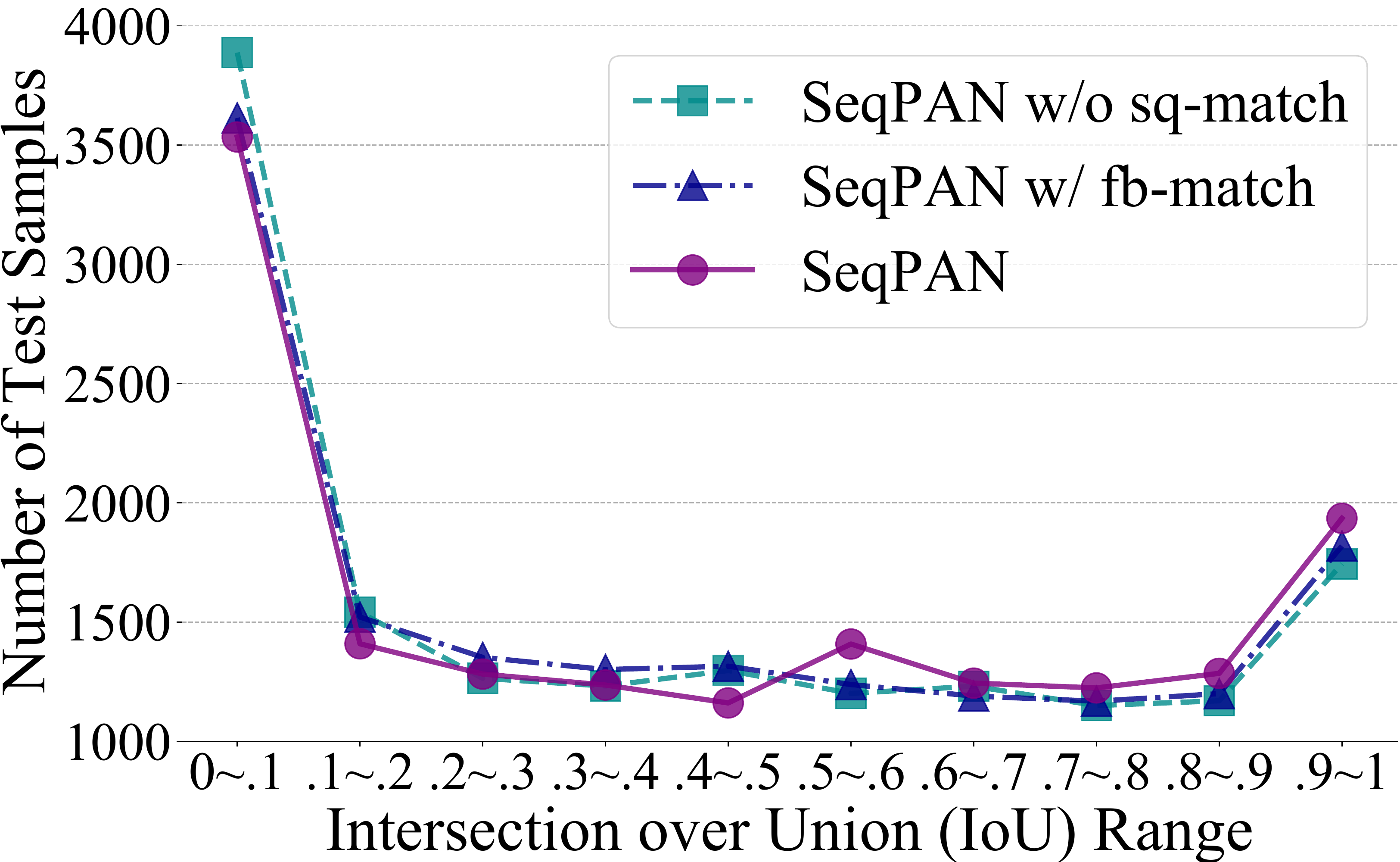}}
	\caption{\small Plots of the number of predicted test samples within different IoU ranges on Charades-STA and ANetCaps.}
	\label{fig:iou_count}
\end{figure}

\subsection{Discussion and Analysis}
We perform in-depth ablation studies to analyze the effectiveness of the SeqPAN. We run all the experiments 5 times and report 5-run average.

\paratitle{Analysis on Self-Guided Parallel Attention.} 
The SGPA (see Figure~\ref{fig:sgpa}) is a variant of transformer (TRM), designed for learning cross-modality interactions between visual and text features. Here, we compare SGPA with standard TRMs. To better reflect the performance of different TRMs, we remove the sequence matching component and only use a single block (\ie $N_{\text{SGPA}}=1$) in this experiment. The results are reported in Table~\ref{tab:transformer}. Observe that SGPA is superior to TRMs on both datasets. Co-TRM performs better on Charades-STA but worse on ANetCaps comparing with Se-TRM. Compared to Se-TRM and Co-TRM, SGPA learns both self-modal contexts and cross-modal interactions, which is approximately equivalent to parallel connection of two modules.

\paratitle{Impact of SGPA block numbers $N_{\text{SGPA}}$.}
We now study the impact of SGPA block numbers on Charades-STA and ANetCaps. We evaluate five different values of $N_{\text{SGPA}}$ from $1$ to $5$. The performance across the number of SGPA blocks in SeqPAN are plotted in Figures~\ref{fig:sgpa_block_charades} and~\ref{fig:sgpa_block_activitynet}. Best performance is achieved at $N_{\text{SGPA}}=2$ on both datasets. In general, along with increasing $N_{\text{SGPA}}$, the performance of SeqPAN first increases and then gradually decreases, on both datasets. We also note that performance on Charades-STA is not very sensitive to the setting of $N_{\text{SGPA}}$.

\paratitle{Analysis on Sequence Matching.}
The conventional matching strategy~\cite{Yuan2019ToFW,lu2019debug,mun2020local} (denoted by fb-match) is to predict whether a frame is inside or outside of target moment, \ie foreground or background. In SeqPAN, we predict begin-, inside- and end-regions, and introduce label embeddings ($\bm{E}_{\text{lab}}$) to represent each region. The prediction process also uses the Gumbel-Max trick. In this experiment, we analyze the effects of label embeddings and Gumbel-Max trick in sequence matching. 

Summarized in Table~\ref{tab:seq_mat}, both Gumbel-Max trick (denoted by $G$) and label embeddings contribute to the grounding performance improvement. In addition, consistent improvements are observed by incorporating $G$ and $\bm{E}_{\text{lab}}$ into the model. SeqPAN is superior to SeqPAN w/ fb-match over all evaluation metrics. The performance improvements are more significant under more strict metrics. The results show that sq-match is more effective than the fb-match strategy. Regional indication of potential positions of start/end boundaries does help the model to produce accurate predictions.

\paratitle{Impact of Annealing Temperature $\tau$.}
We then analyze the impact of annealing temperature $\tau$ of Gumbel-Softmax in sequence matching module. Gumbel-Softmax distributions are identical to a categorical distribution when $\tau\to 0^{+}$. With $\tau\to\infty$, its distribution is smooth. We evaluate 11 different $\tau$ values from $0.01$ to $1.0$, where $0.01$ is used to approximate $0.0$ since $0.0$ is not divisible. The results are compared against vanilla Softmax as a baseline. For vanilla Softmax, we multiply the probability distribution of labels with $\bm{E}_{\text{lab}}$, to aggregate label information into the visual representations.

Figure~\ref{fig:annealing_temperature} plots the results of different $\tau$'s on Charades-STA and ANetCaps, respectively. We observe similar patterns on the four sets of results. The best performance is achieved when $\tau=0.3$ over both metrics on both datasets. From Figure~\ref{fig:tau_charades}, when $\tau$ is too small or too large (\ie the probability distribution from Gumbel-Softmax becomes too sharp or too smooth), Gumbel-Softmax performs poorer than vanilla Softmax. This result suggests that a proper annealing temperature $\tau$ is crucial to achieve good performance. Similar observations hold on ANetCaps (see Figure~\ref{fig:tau_activitynet}).

\begin{figure}[t]
    \centering
    \includegraphics[width=0.47\textwidth]{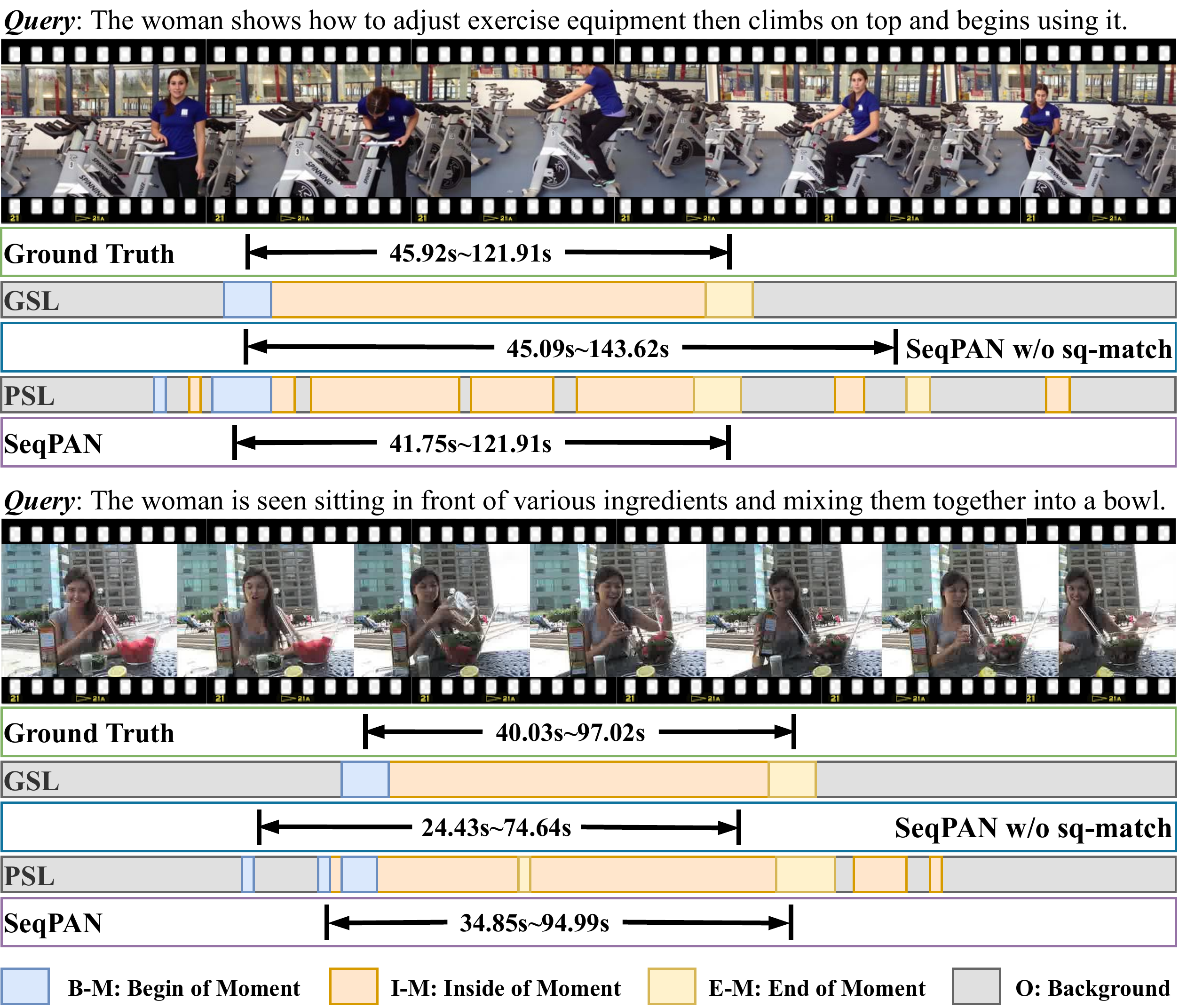}
    \caption{\small Qualitative results of SeqPAN and SeqPAN w/o sq-match on ANetCaps. ``GSL'' is the ground truth sequence labels; ``PSL'' is the predicted labels by sq-match of SeqPAN.}
	\label{fig:visualization}
\end{figure}

\subsection{Qualitative Analysis}
Figure~\ref{fig:iou_count} shows the number of predicted test samples within different IoU ranges on Charades-STA and ANetCaps. 
Here, we compare SeqPAN with two of its variants: (i) removal of sequence matching module, and (ii) replacement of sequence matching with fb-match. All three variants show similar patterns. Nevertheless, within the higher IoU ranges, \eg $\text{IoU}\geq 0.5$ on both datasets,  SeqPAN and the variant with fb-match outperform the variant without sequence matching. The results show that having auxiliary objectives (\eg foreground/background or sequential regions) is helpful in video grounding task. Results in Figure~\ref{fig:iou_count} also show that our sequence matching is more effective than fb-match, for highlighting the correction regions for predicting start/end boundaries.  

Figure~\ref{fig:visualization} depicts two video grounding examples from the ANetCaps dataset. From the two examples, the moments retrieved by SeqPAN are closer to the ground truth than that are retrieved by SeqPAN without utilizing the sq-match strategy. Besides, the start and end boundaries predicted by SeqPAN are roughly constrained within the pre-set potential start and end regions. In addition, the predicted sequence labels (PSL) in Figure~\ref{fig:visualization} also reveal the weakness of sq-match strategy. The predicted labels by sq-match strategy are not continuous, where multiple start, inside, and end regions are generated. In consequence, the localizer may be affected by wrongly predicted regions and leads to inaccurate results. To further constrain the generated regions is part of our future work.  

\section{Conclusion}
In this work, we propose a Parallel Attention Network with Sequence matching (SeqPAN) to address the language query-based video grounding problem. We design a parallel attention module to improve the multimodal representation learning by capturing both self- and cross-modal attentive information simultaneously. In addition, we propose a sequence matching strategy, which explicitly indicates the potential start and end regions of the target moment to allow the localizer precisely predicting the boundaries. Through extensive experimental studies, we show that SeqPAN outperforms the state-of-the-art methods on three benchmark datasets; and both the proposed parallel attention and sequence matching modules contribute to the grounding performance improvement.

\section*{Acknowledgments}
This research is supported by the Agency for Science, Technology and Research (A*STAR) under its AME Programmatic Fund (Project No. A18A1b0045 and A18A2b0046).

\bibliographystyle{acl_natbib}
\bibliography{acl2021}

\clearpage

\appendix

\noindent This appendix contains two sections. Section~\ref{apd:sec:model} provides (\ref{apd:sec:sgpa}) a detailed comparison between the proposed SGPA and standard transformer blocks, (\ref{apd:sec:vqi})  technical details of the video-query integration module,  and (\ref{apd:sec:gumbel}) categorical reparameterization used in the sequence matching module. Section~\ref{apd:sec:settings} describes statistics on the benchmark datasets and parameter settings in our experiments.

\section{Additional Comparison and Technical Details}\label{apd:sec:model}

\subsection{SGPA versus Standard Transformers}\label{apd:sec:sgpa}
\begin{figure*}[t]
    \centering
	\subfigure[\small Transformer Block (Self Attn)]
	{\label{fig:se_trm}	\includegraphics[width=0.32\textwidth]{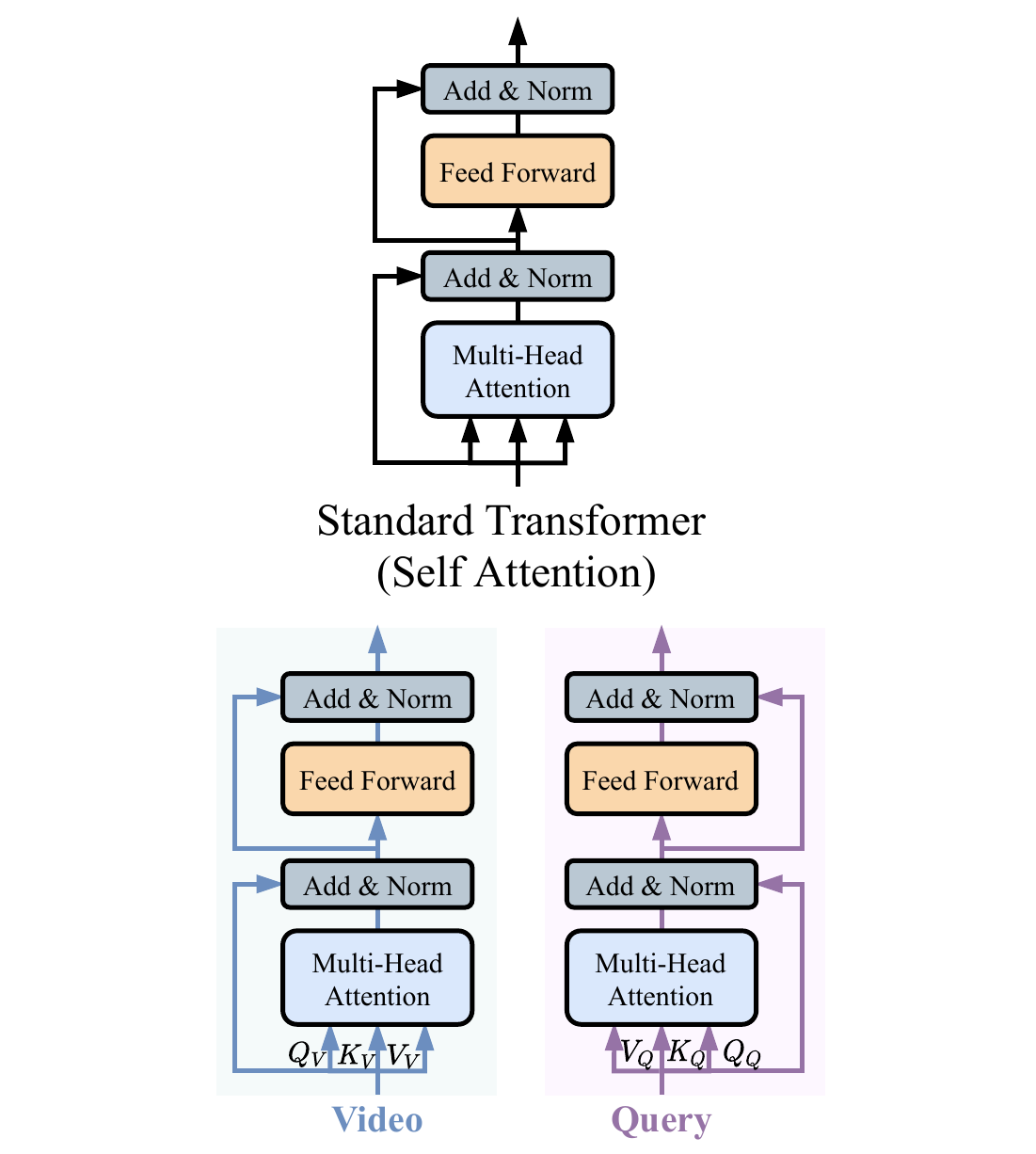}}
	\subfigure[\small Transformer Block (Cross Attn)]
	{\label{fig:co_trm}	\includegraphics[width=0.32\textwidth]{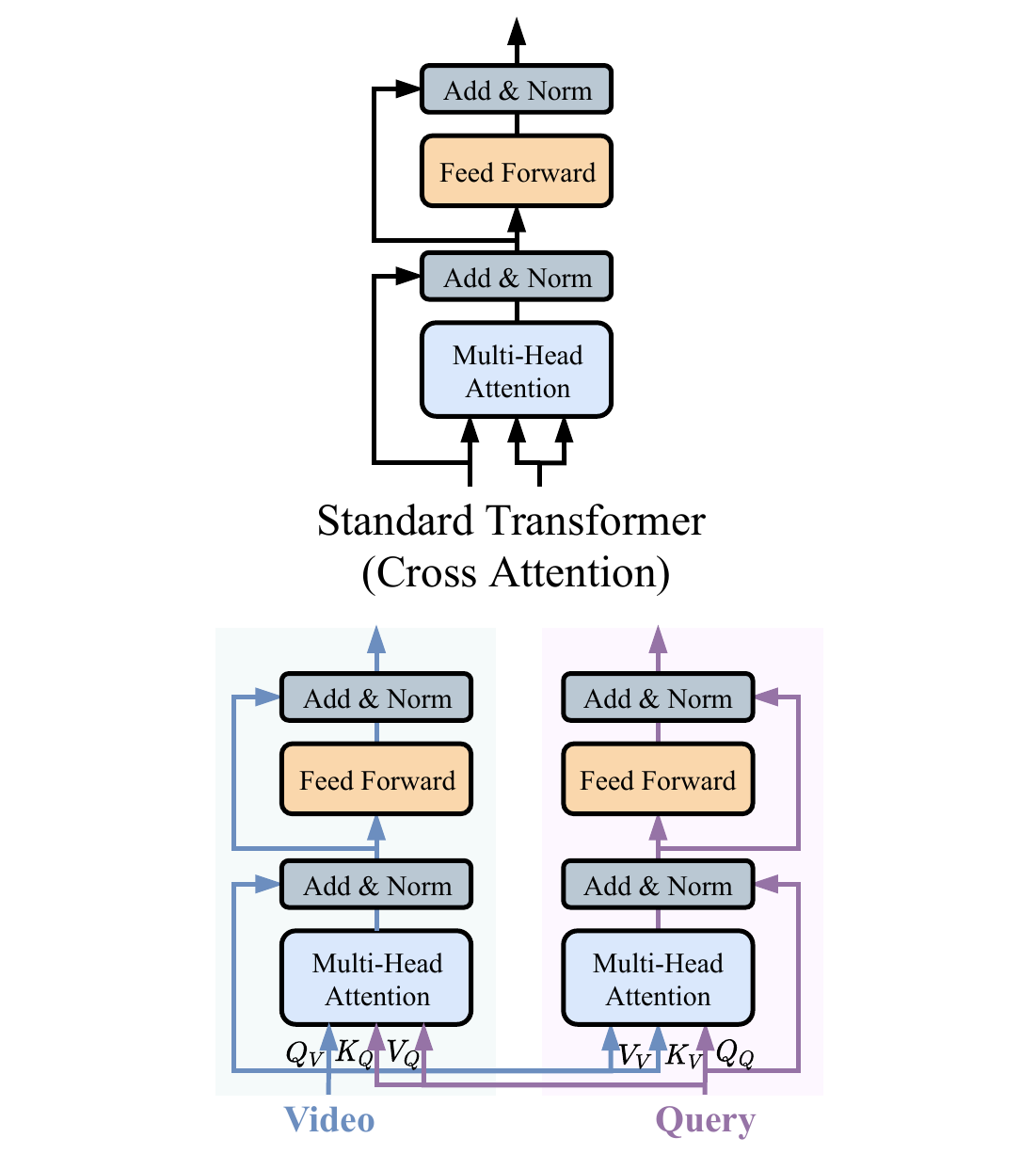}}
	\subfigure[\small Self Guided Parallel Attention (Ours)]
	{\label{fig:sgpa_trm}\includegraphics[width=0.32\textwidth]{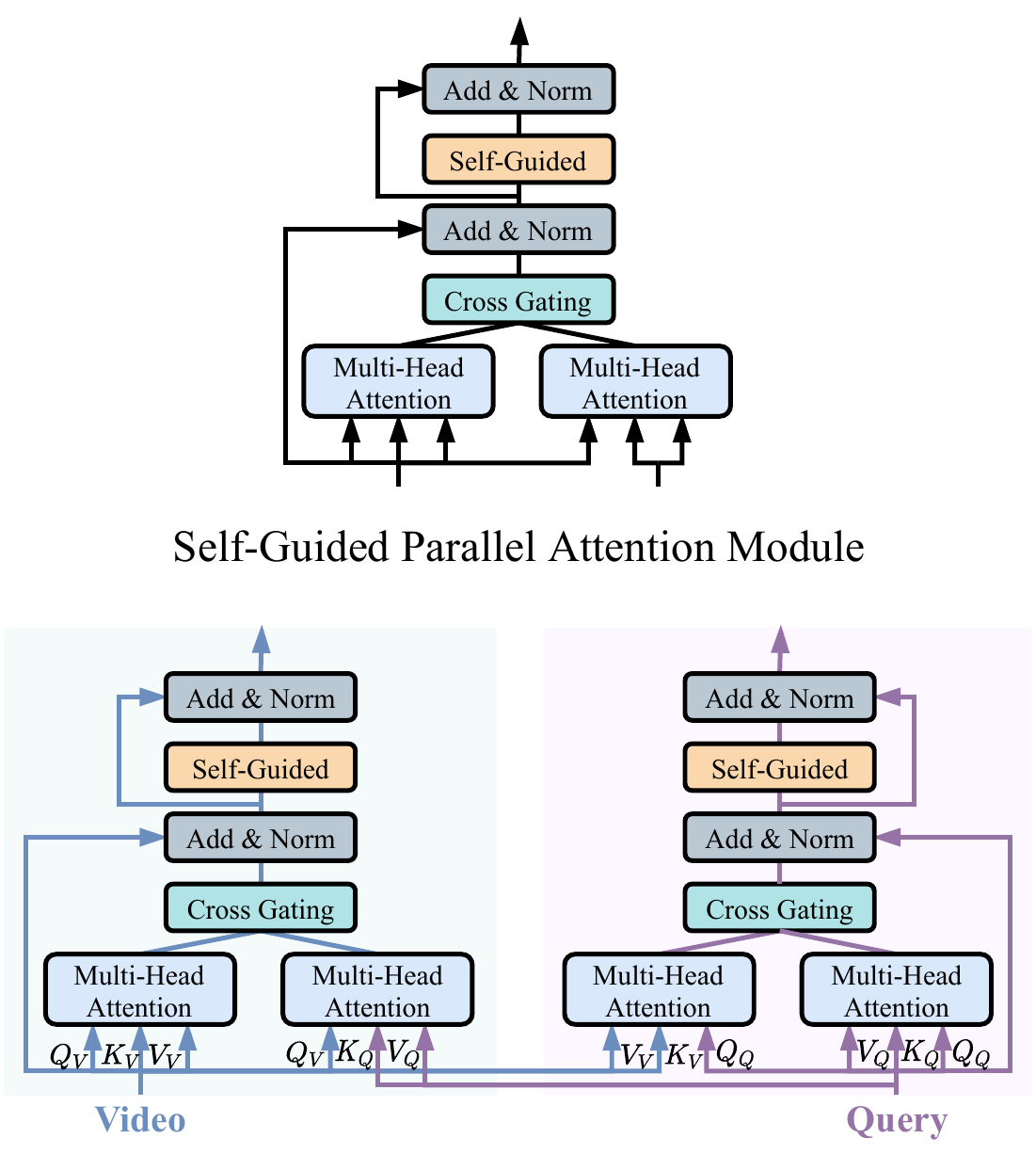}}
	\vspace{-3mm}
	\caption{\small The structures of standard transformer blocks and self-guided parallel attention module. Top: the structure of each module; Bottom: the parallel streams of encoding visual and textual inputs. (a) The standard transformer block with self-attention; (b) The standard transformer block with cross-attention; (c) The self-guided parallel attention (SGPA) module.}
	\label{fig:transformer}
\end{figure*}

Two ways are mainly used to adopt the transformer block for multi-modal representation learning: 
\begin{itemize}[itemsep=-0.3ex,leftmargin=2.5ex]
    \item Transformer block with the self-attention (Se-TRM), which encodes visual and textual inputs in separate streams, shown in Figure~\ref{fig:se_trm}.
    \item Transformer block with the cross-attention (Co-TRM), which encodes both visual and textual inputs with interactions through co-attention, shown in  Figure~\ref{fig:co_trm}.
\end{itemize}

Several works~\cite{lu2019debug,chen2020rethinking,zhang2020vslnet} adopt Se-TRM to learn visual and textual representations in video grounding task. Se-TRM separately encodes each modality, it focuses on learning the refined unimodal representations within each modality for video and text respectively. Without any connection between two modalities, Se-TRM cannot use information from other modality to improve the representations. 

Co-TRM\footnote{It is also known as co-attentional, multi-modal or cross-modal transformer block in different works.} is commonly used as a basic component in various vision-language methods~\cite{tan2019lxmert,lu2019vilbert,lei2020tvr}. Co-TRM relies on co-attention to learn the cross-modal representations for both visual and textual inputs. However, Co-TRM lacks the ability to encode self-attentive context within each modality.

\begin{figure}[t]
    \centering
	\subfigure[\small SeCo-TRM Block]
	{
	\label{fig:seco_trm}	
	\includegraphics[trim={0mm 11mm 8mm 2mm},clip,width=0.22\textwidth]{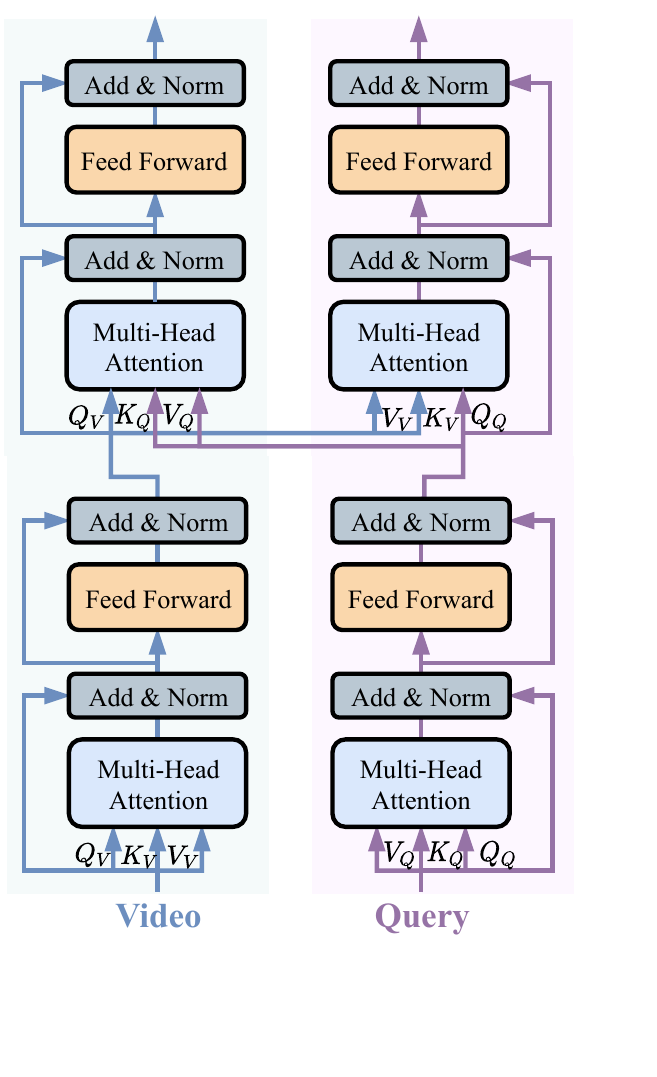}
	}
	\subfigure[\small CoSe-TRM Block]
	{
	\label{fig:cose_trm}	
	\includegraphics[trim={0mm 11mm 8mm 2mm},clip,width=0.22\textwidth]{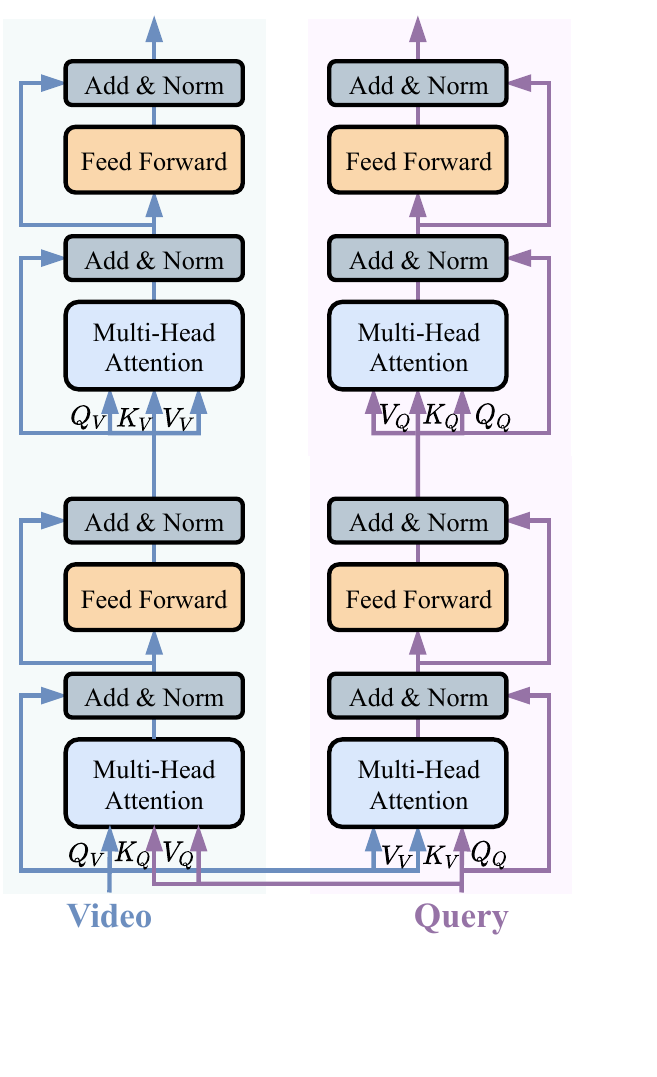}
	}
	\vspace{-3mm}
	\caption{\small The structures of SeCo-TRM and CoSe-TRM.}
	\label{fig:stack_transformer}
\end{figure}

The cascade of Se-TRM and Co-TRM is also used in recent vision-language models~\cite{tan2019lxmert,lu2019vilbert,zhu2020actbert,lei2020tvr} to learn both unimodal and cross-modal representations. In general, there are two cascade forms: 1) stacking Co-TRM upon Se-TRM (SeCo-TRM) in Figure~\ref{fig:seco_trm}; and 2) stacking Se-TRM upon Co-TRM (CoSe-TRM) in Figure~\ref{fig:cose_trm}. These stacked TRMs learn the unimodal and cross-modal information in a sequence manner. Hence, their final outputs focus more on either the self-attentive contexts or cross-modal interactions. 
Our SGPA combines advantages of both Se-TRM and Co-TRM, but not through cascade. As shown in Figure~\ref{fig:sgpa_trm}, SGPA contains two parallel multi-head attention blocks. One block takes single modality as input and the other takes both modalities as inputs. Thus, SGPA is able to learn both unimodal and cross-modal representations simultaneously. Then, a cross-gating strategy is designed to fuse the self- and cross-attentive representations. We also employ a self-guided head to replace the feed forward layer in transformer block. This design implicitly emphasizes informative representations by measuring the confidence of each element.

Table~\ref{tab:stack_transformer} reports the performance of SGPA and standard TRMs on Charades-STA and ANetCaps datasets. Here, we regard both SeCo-TRM and CoSe-TRM as single block. The results show that both PA (a SGPA variant without self-guided head) and SGPA are superior to standard TRMs.

\begin{table}[t]
    \small
	\centering
	\begin{tabular}{l | c c c}
		\specialrule{.1em}{.05em}{.05em}
		\multirow{2}{*}{Methods} & \multicolumn{3}{c}{$\text{R@}1, \text{IoU}=\mu$} \\
        & $\mu=0.3$ & $\mu=0.5$ & $\mu=0.7$ \\
        \hline
        \multicolumn{4}{c}{Charades-STA} \\
        \hline
        Se-TRM & 68.84~\scriptsize{(0.46)} & 51.92~\scriptsize{(0.54)} & 34.58~\scriptsize{(0.18)} \\
        Co-TRM & 69.03~\scriptsize{(0.49)} & 52.34~\scriptsize{(0.50)} & 35.07~\scriptsize{(0.32)} \\
        SeCo-TRM & 69.11~\scriptsize{(0.24)} & 52.63~\scriptsize{(0.49)} & 35.17~\scriptsize{(0.22)} \\
        CoSe-TRM & 69.08~\scriptsize{(0.26)} & 52.82~\scriptsize{(0.43)} & 35.09~\scriptsize{(0.50)} \\
        PA     & 69.21~\scriptsize{(0.27)} & 54.37~\scriptsize{(0.46)} & 36.22~\scriptsize{(0.49)} \\
        SGPA   & 69.47~\scriptsize{(0.32)} & 54.63~\scriptsize{(0.43)} & 36.36~\scriptsize{(0.24)} \\
        \hline
        \multicolumn{4}{c}{ActivityNet Captions} \\
        \hline
        Se-TRM & 57.64~\scriptsize{(0.38)} & 40.76~\scriptsize{(0.35)} & 25.10~\scriptsize{(0.30)} \\
        Co-TRM & 57.39~\scriptsize{(0.29)} & 40.55~\scriptsize{(0.45)} & 24.85~\scriptsize{(0.47)} \\
        SeCo-TRM & 57.47~\scriptsize{(0.38)} & 40.70~\scriptsize{(0.24)} & 25.07~\scriptsize{(0.21)} \\
        CoSe-TRM & 57.72~\scriptsize{(0.41)} & 40.85~\scriptsize{(0.17)} & 25.16~\scriptsize{(0.15)} \\
        PA     & 58.27~\scriptsize{(0.13)} & 41.59~\scriptsize{(0.24)} & 25.88~\scriptsize{(0.28)} \\
        SGPA   & 58.40~\scriptsize{(0.31)} & 41.72~\scriptsize{(0.19)} & 26.07~\scriptsize{(0.16)} \\
        \specialrule{.1em}{.05em}{.05em}
	\end{tabular}
	\caption{\small Comparison between SGPA with standard transformer blocks on Charades-STA and ANetCaps, where PA is the SGPA without self-guided head (\ie replaced by FFN) The scores in bracket denotes standard deviation.}
	\label{tab:stack_transformer}
\end{table}

\subsection{Video-Query Integration Computation}\label{apd:sec:vqi}
This section presents the detailed computation process of video-query integration (see Section 3.1.3). 

Given two inputs $\bm{X}\in\mathbb{R}^{d\times N_x}$ and $\bm{Y}\in\mathbb{R}^{d\times N_y}$, the context-query attention first computes similarities between each pair of $\bm{X}$ and $\bm{Y}$ elements as:
\begin{equation}
    \mathcal{S}=\bm{X}^{\top}\cdot\bm{W}\cdot\bm{Y}
\end{equation}
where $\bm{W}\in\mathbb{R}^{d\times d}$ and $\mathcal{S}\in\mathbb{R}^{N_x\times N_y}$. Then $X$-to-$Y$ and $Y$-to-$X$ attention weights are computed by:
\begin{equation}
\begin{aligned}
    \mathcal{A}_{\text{XY}} & = \mathcal{S}_r \cdot \bm{Y}^{\top}\in\mathbb{R}^{N_x\times d} \\
    \mathcal{A}_{\text{YX}} & = \mathcal{S}_r \cdot \mathcal{S}^{\top}_c \cdot \bm{X}^{\top}\in\mathbb{R}^{N_x\times d}
\end{aligned}
\end{equation}
where $\mathcal{S}_{r}$ and $\mathcal{S}_{c}$ are the row- and column-wise normalization of $\mathcal{S}$ by Softmax function. The final output of context-query attention is calculated as:
\begin{equation}
    \bm{X}^{Y}=\mathtt{FFN}\big([\bm{X};\mathcal{A}_{\text{XY}}^{\top};\bm{X}\odot\mathcal{A}_{\text{XY}}^{\top};\bm{X}\odot\mathcal{A}_{\text{YX}}^{\top}]\big)
\end{equation}
where $\odot$ denotes element-wise multiplication, ``$;$'' represents concatenation operation, and $\bm{X}^{Y}\in\mathbb{R}^{d\times N_x}$. In this way, the information of $\bm{Y}$ is properly fused into $\bm{X}$.

By setting $\bm{X}=\bm{\bar{V}}\in\mathbb{R}^{d\times N}$ and $\bm{Y}=\bm{\bar{Q}}\in\mathbb{R}^{d\times M}$, we can derive the query-aware video representations $\bm{V}^{Q}\in\mathbb{R}^{d\times N}$. Similarly, the video-aware query representations $\bm{Q}^{V}\in\mathbb{R}^{d\times M}$ is obtained by setting $\bm{X}=\bm{\bar{Q}}$ and $\bm{Y}=\bm{\bar{V}}$. 

Next, we encode $\bm{Q}^{V}=[\bm{q}_{0}^{V},\dots,\bm{q}_{M-1}^{V}]$ into sentence representation $\bm{q}$ with additive attention:
\begin{equation}
\begin{aligned}
    \bm{\alpha} & = \mathtt{Softmax}\big(\bm{W}_{\alpha}\cdot\bm{Q}^{V})\big) \in \mathbb{R}^{M} \\
    \bm{q} & = \sum_{i=0}^{M-1}\alpha_i\times\bm{q}_i^V \in \mathbb{R}^{d}
\end{aligned}
\end{equation}
where $\bm{W}_{\alpha}\in\mathbb{R}^{1\times d}$. The $\bm{q}$ is then concatenated with each element of $\bm{V}^{Q}$ as $\bm{H}=[\bm{h}_{1},\dots,\bm{h}_{n}]\in\mathbb{R}^{2d\times N}$, where $\bm{h}_{i}=[\bm{v}_{i}^{Q};\bm{q}]$. Finally, the query-attended visual representation is computed as
\begin{equation}
    \bm{\bar{H}}=\bm{W}_h\cdot\bm{H}+\bm{b}_h
\end{equation}
where $W_h\in\mathbb{R}^{d\times 2d}$ and $b_h\in\mathbb{R}^{d}$ denote the learnable weight and bias, and $\bm{\bar{H}}\in\mathbb{R}^{d\times N}$.

\begin{table*}[t]
    \small
    \setlength{\tabcolsep}{2.7pt}
	\centering
	\begin{tabular}{l c c c r r r r r}
		\toprule
		Dataset & Domain & $N_{\text{V}}$ (train/val/test) & $N_{\text{A}}$ (train/val/test) & $\bar{N}_{\text{A/V}}$ & $N_{\text{Vocab}}$ & $\bar{L}_{V}$ (s) & $\bar{L}_{Q}$ & $\bar{L}_{M}$ (s) \\
		\midrule
        Charades-STA & Indoors & $5,338/\text{-}/1,334$ & $12,408/\text{-}/3,720$ & $2.42$ & $1,303$ & $30.59$ & $7.22$ & $8.22$ \\
        \midrule
        ActivityNet Captions & Open & $10,009/\text{-}/4,917$ & $37,421/\text{-}/17,505$ & $3.68$ & $12,460$ & $117.61$ & $14.78$ & $36.18$ \\
        \midrule
        TACoS~\cite{Gao2017TALLTA} & \multirow{2}{*}{Cook} & \multirow{2}{*}{$75/27/25$} & $10,146/4,589/4,083$ & $148.17$ & $2,033$ & $287.14$ & $10.05$ & $5.45$ \\
        TACoS~\cite{zhang2020learning} & & & $9,790/4,436/4,001$ & $143.52$ & $1,983$ & $287.14$ & $9.42$ & $25.26$ \\
        \bottomrule
	\end{tabular}
    \caption{\small Statistics of the evaluated video grounding benchmark datasets, where $N_{\text{V}}$ is number of videos, $N_{\text{A}}$ is number of annotations, $\bar{N}_{\text{A/V}}$ denotes the average number of annotations per video, $N_{\textrm{Vocab}}$ is the vocabulary size of lowercase words, $\bar{L}_{V}$ denotes the average length of videos in seconds, $\bar{L}_{Q}$ denotes the average number of words in the sentence queries and $\bar{L}_{M}$ represents the average length of temporal moments in seconds.}
	\label{tab-data}
\end{table*}

\subsection{Categorical Reparameterization}\label{apd:sec:gumbel}
This section provides a brief introduction of the categorical reparameterization strategy used in sequence matching module (see Section 3.1.4).

Categorical reparameterization, \eg reinforce-based approaches~\cite{sutton2000policy,schulman2015gradient}, straight-through estimators~\cite{bengio2013estimating} and Gumbel-Softmax~\cite{jang2017categorical,maddison2017the}, is a strategy that enables discrete categorical variables to back-propagate in neural networks. It aims to estimate smooth gradient with a continuous relaxation for categorical variable. In this work, we use Gumbel-Softmax to approximate the sequence labels from a probability distribution. Then those labels are applied to lookup the corresponding embeddings for region representation in the sequence matching module of SeqPAN. 

Let $\bm{x}=(x_1,\dots,x_l)$ be a categorical distribution, where $l$ is the number of categories, $x_c$ is the probability score of category $c$ and $\sum_{c=1}^{l}x_c=1$. Given the \textit{i.i.d.} Gumbel noise $\bm{g}=(g_1,\dots,g_l)$ from $\mathtt{Gumbel}(0,1)$ distribution\footnote{The $\mathtt{Gumbel}(0,1)$ distribution can be sampled using inverse transform sampling by drawing $u\sim\mathtt{Uniform}(0,1)$ and computing $g=-\log(-\log(u))$~\cite{jang2017categorical}.}, the soft categorical sample can be computed as:
\begin{equation}
    \bm{y}=\mathtt{Softmax}\big((\log(\bm{x})+\bm{g})/\tau\big)
\label{eq:gumbel_softmax_appd}
\end{equation}
where $\tau>0$ is annealing temperature, and Eq.~\ref{eq:gumbel_softmax_appd} is referred as Gumbel-Softmax operation on $\bm{x}$. As $\tau\to 0^{+}$, $\bm{y}$ is equivalent to the Gumbel-Max form~\cite{gumbel1954statistical,maddison2014asampling} as:
\begin{equation}
    \bm{\hat{y}}=\mathtt{Onehot}\big(\arg\max(\log(\bm{x})+\bm{g})\big)
\label{eq:gumbel_max_appd}
\end{equation}
where $\bm{\hat{y}}$ is an unbiased sample from $\bm{x}$ and thus we can draw differentiable samples from the distribution during training. Note, when input $\bm{x}$ is unnormalized, the $\log(\cdot)$ operator in Eq.~\ref{eq:gumbel_softmax_appd} and~\ref{eq:gumbel_max_appd} shall be omitted~\cite{jang2017categorical,dong2019searching}. During inference, discrete samples can be drawn with the Gumbel-Max trick directly.

\section{Dataset and Parameter Settings}\label{apd:sec:settings}

\subsection{Dataset Statistics}\label{apd:sec:dataset}
The statistics of the evaluated benchmark datasets are summarized in Table~\ref{tab-data}. \textbf{Charades-STA} dataset consists of $6,672$ videos and $16,128$ annotations (\ie moment-query pairs) in total. \textbf{ActivityNet Captions (ANetCaps)} dataset is taken from the ActivityNet~\cite{heilbron2015activitynet}. The average duration is about $120$ seconds and each video contains $3.68$ annotations on average. \textbf{TACoS} dataset contains $127$ cooking activities videos with average duration of $4.79$ minutes, and $18,818$ annotations in total. We follow the same train/val/test split as \citet{Gao2017TALLTA}. Besides, \citet{zhang2020learning} pre-processes the TACoS dataset, hence their version is slightly different from the original version. Detailed statistics are summarized in Table~\ref{tab-data}.

\subsection{Hyper-Parameter Settings}\label{apd:sec:hyperparam}
We follow~\cite{ghosh2019excl,mun2020local,rodriguez2020proposal,zhang2020vslnet} and use 3D ConvNet pre-trained on Kinetics dataset (\ie I3D\footnote{https://github.com/deepmind/kinetics-i3d})~\cite{Carreira2017QuoVA} to extract visual features from videos. The maximal visual feature sequence lengths are set to $64$, $100$, and $256$ for Charades-STA, ActivityNet Captions, and TACoS, respectively. This setting is based on the average video lengths in the three datasets. The feature sequence length of a video will be uniformly downsampled if it is larger than the pre-set threshold, or zero-padding otherwise. For the language queries, we lowercase all the words and initialize them with GloVe~\cite{pennington2014glove} embeddings\footnote{http://nlp.stanford.edu/data/glove.840B.300d.zip}. The word embeddings and extracted visual features are fixed during training.

For other hyper-parameters, we use the same settings for all datasets. The dimension of the hidden layers is $128$; the head number in multi-head attention is $8$; the number of SGPA blocks ($N_{\text{SGPA}}$) is $2$; the annealing temperature $\tau$ of Gumbel-Softmax is $0.3$; The Dropout~\cite{srivastava2014dropout} is $0.2$. The maximal training epochs $E=100$ is used, with batch size of $16$ and early stopping tolerance of $10$ epochs. We adopt Adam~\cite{Kingma2015AdamAM} optimizer, with initial learning rate of $\beta_0=0.0001$, weight decay $0.01$, and gradient clipping $1.0$, to train the model. The learning rate decay strategy is defined as $\beta_{e}=\beta_0\times(1-\frac{e}{E})$, where $e$ denotes the $e$-th training epoch.

The SeqPAN is implemented using TensorFlow $\texttt{1.15.0}$ with CUDA $\texttt{10.0}$ and cudnn $\texttt{7.6.5}$. All the experiments are conducted on a workstation with dual NVIDIA GeForce RTX 2080Ti GPUs.

\end{document}